\documentclass[11pt]{article}
\usepackage{fontspec}
\usepackage[margin=1in]{geometry}
\usepackage{amsmath,amssymb,graphicx,booktabs,placeins,listings,array}
\usepackage{natbib}
\usepackage[hidelinks]{hyperref}
\hypersetup{
  pdftitle={Local Budget Benefits, Population Commitment, and Solver Performance: A DGO Case Study with Controlled DE Transfer},
  pdfauthor={Shiqi Li},
  pdfsubject={Evaluation contracts for population-based optimization},
  pdfkeywords={population optimization, differential evolution, evaluation budget, counterfactual replay, interpolation}}
\title{Local Budget Benefits, Population Commitment, and Solver Performance:\\A DGO Case Study with Controlled DE Transfer}
\author{Shiqi Li\\{\small School of Traffic and Transportation,}\\{\small Lanzhou Jiaotong University}}
\date{}
\begin{document}
\maketitle

\begin{abstract}
An evaluated candidate can improve a local search step, alter the population used by later steps, and affect a solver's final result. We distinguish these three questions using Da Gou Gun Optimization (DGO), a population optimizer designed for this study. Its JQ100 configuration evaluates an interpolation candidate during a two-sided probe. We define local budget value, population commitment effects and complete-policy performance, and give an exact example in which a locally better commitment leads to a worse continuation. Short-horizon matched comparisons favor interpolation on eligible quadratic states. Paired continuations yield mixed effect estimates, with primary group intervals spanning zero. Complete-policy comparisons favor JQ100 over the tested classical references in aggregate; a matched-start comparison has favorable means but intervals spanning parity. On 24 boxed BBOB functions, target-truncated geometric-mean error ratios favor JQ100 over the tested classical policies and favor L-SHADE and CMA-ES over JQ100 under the stated initialization and stopping rules. Local evaluation gains, commitment effects and solver competitiveness therefore require separate comparisons.
\end{abstract}

\section{Introduction}
Evaluating a candidate provides information; writing it into the population changes future search. We call this population update \emph{commitment}. The value of the evaluation, the effect of commitment and the performance of the complete solver need not have the same ordering. A terminal score alone cannot separate them.

We study this distinction with Da Gou Gun Optimization (DGO), a modular population optimizer developed for this research. Its fixed JQ100 configuration adds an eligible interpolation point to a two-sided probe. Earlier quadratic tests showed aggregate gains alongside a severe individual loss, motivating a closer examination of what happens after a useful local evaluation. Our KEEP/DROP intervention changes one population slot while retaining the evaluated point, its contribution to historical best and its cost.

Operator-credit studies and controlled evolutionary replays examine delayed outcomes \citep{whitacre2006,siper2026,ferguson2026}. Here we hold the paid information fixed and change only a candidate's population commitment. An exact construction shows how this change can reverse a local ordering. Matched local tests, DGO and DE-plus-probe continuations, and complete-policy comparisons then examine the distinction empirically.

The contribution is a set of explicit comparison quantities, a paid-history-preserving intervention, and mathematical conditions linking commitment to policy performance. The empirical argument proceeds through four steps: basic optimization effectiveness, local budget value, population commitment, and competition with stronger solvers. The first and last compare complete policies; the middle steps examine decisions within search.

\section{Related work}
Two-sided directional search and model-based derivative-free methods provide established frameworks for spending evaluations \citep{bergou2019stp,conn2009dfo}; quadratic interpolation has also been combined with differential evolution \citep{noman2008local,tardivo2012}. DGO's reflected, quota-controlled probe has its own eligibility and accounting rules. The broader question of how to allocate evaluations also appears in Bayesian optimization \citep{jones1998ego,astudillo2021multistep}.

Population variation and memory shape future search. Differential evolution uses population differences \citep{storn1997de}, while adaptive and archive-based methods change parameters, population size, or donor history \citep{zhang2009jade,tanabe2014lshade}. Analyses and algorithms have addressed diversity, strategy choice and historical selection \citep{zaharie2017variance,meng2024acd,feng2026diversity,kitamura2026selection}. Operator-selection studies distinguish immediate from delayed credit \citep{fialho2010bandits,karafotias2015control,whitacre2006}. Controlled evolutionary continuations and population replays are close precedents \citep{siper2026,ferguson2026}; here the paid observation is held fixed while one commitment changes.

Complete-solver comparisons also depend on initialization, stopping and budgets. Performance profiles and COCO support problem- and target-based benchmarking \citep{dolan2002profiles,hansen2021coco,hansen2022anytime}; a prefix of a long-budget run need not represent a separately configured short-budget policy \citep{tusar2017budget}. We compare specified classical policies and stronger L-SHADE and CMA-ES references under stated initialization and stopping rules \citep{tanabe2014lshade,hansen2023cma,clerc2002particle}.

\section{Methods}
\subsection{DGO and accounting}
We minimize a deterministic objective on a finite box. Each evaluated point costs one function evaluation (FE), including initialization, rejected candidates and repeated points; historical best is the minimum over all paid observations. JQ100 uses 40 individuals, a 0.5 sweep/probe evaluation quota, pair-based radius feedback and no scout. Each probe evaluates two reflected endpoints and, when eligible and budgeted, an interpolated point. The best strict improvement is committed; endpoint-pair outcomes alone update the radius. NQ disables interpolation, RQ100 uses a random interior third point, and S0 uses pure sweep.

JQ100's initial radius is $0.1\sqrt D$ in unit-box coordinates, its radius-feedback parameter is $\eta=0.7$, and the elite fraction is 20\%. The name JQ100 denotes evaluation of every eligible third point allowed by the remaining budget, not an extra budget allowance. The historical-archive configuration H10 is a separate policy and appears only as supporting evidence.

\begin{figure}[!htbp]
\centering
\includegraphics[width=\linewidth]{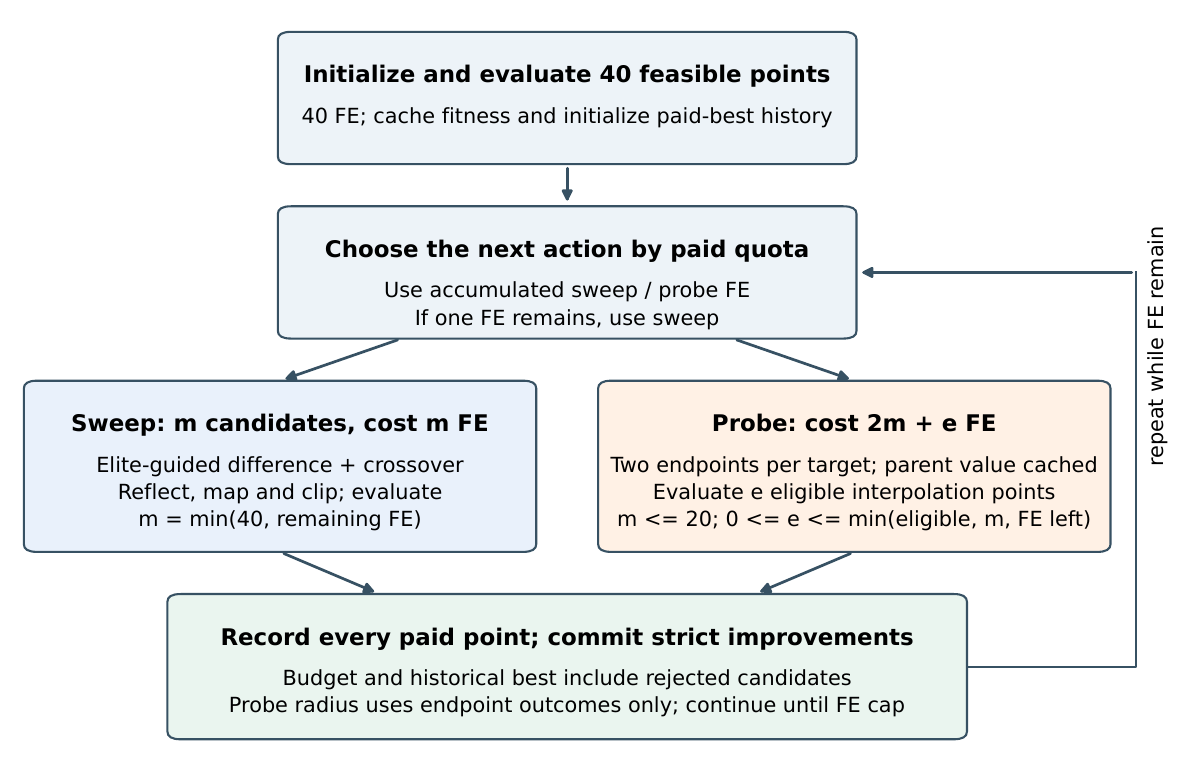}
\caption{Core workflow of the fixed JQ100 quota policy. Initialization, sweep trials, probe endpoints and eligible interpolation points consume evaluations; every paid value updates historical best. Endpoint-pair outcomes update the probe radius.}
\label{fig:dgo-flow}
\end{figure}

For target $i$, the sweep forms
\[
v_i=x_i+F_i[(x_\ell-x_i)+(x_{r_1}-x_{r_2})],\qquad F_i\sim U[0.4,0.9].
\]
Trials use the population at the start of the batch, an elite $\ell$ and two distinct eligible donors. Binomial crossover, reflection and strict selection complete the sweep. Because later trials use population coordinates, changing one slot can change future candidates even when donor indices are held fixed. Replays therefore preserve the population, values, radii, history, counters and random-generator state. Figure~\ref{fig:dgo-flow} and the outline below show the main steps; Appendix E gives the full algorithms.

\noindent\begin{minipage}{\linewidth}
\begin{lstlisting}[caption={Outline of the fixed quota probe--sweep policy.}]
Initialize population, radii, counters and historical best
Evaluate the initial population and charge every point
WHILE evaluation budget remains:
    Select sweep or probe from their accumulated FE quotas
    IF probe is selected and at least two FE remain:
        Select target slots and sample directions
        Evaluate reflected endpoint pairs and charge all points
        Determine eligible interpolation points and remaining cap
        IF at least one point is eligible and budgeted:
            Evaluate the target-order prefix and charge all points
        Select the best strictly improving parent, endpoint or point
        Commit selected positions simultaneously
        Update radii from endpoint-pair success alone
    ELSE:
        Draw elite-guided difference trials from the batch-start state
        Evaluate the budgeted trials and charge all points
        Commit strictly improving trials simultaneously
    Update allocation counters, scores and clocks
    Update historical best from every paid observation
RETURN the best of all paid observations
\end{lstlisting}
\end{minipage}

\subsection{A well-defined finite-budget optimizer}\label{sec:validity}
\textbf{Proposition (feasibility, budget completion and historical-best monotonicity).} Let the population size satisfy $N\ge4$ and the integer budget satisfy $B\ge N$. Assume valid finite box bounds, a feasible initial population, admissible quota-policy parameters, finite candidate arithmetic and finite objective returns, and no external interruption or evaluation failure. Every evaluated point lies in the box; a normally completed run consumes exactly $B$ objective-point evaluations, including initialization; and its historical best never increases.

\textit{Proof.} The evaluator maps candidates to physical coordinates and clips them to the box before calling the objective. Initialization consumes $N$ evaluations. At remaining budget $r>0$, a sweep evaluates $\min(N,r)$ candidates. A probe is used only when $r\ge2$ and evaluates $2\min(\lfloor N/2\rfloor,\lfloor r/2\rfloor)$ endpoints; its optional third-point batch is capped by the budget remaining after those endpoints. Thus every loop iteration consumes at least one and at most $r$ evaluations. The integer remainder reaches zero after at most $B-N$ nonempty loop iterations. Every evaluated batch updates the historical best by taking the minimum of the previous best and the newly observed values, irrespective of population retention. This proves the three properties. The statement concerns logical FE within a normal solver run; separately recorded validation and replay costs are not included in $B$.

The proposition establishes a valid finite-budget search process. Comparing its optimization quality with another trajectory requires a separate criterion, defined next and tested against Random, DE and PSO.

\subsection{Three comparison contracts}
The expectations below assume measurable transitions and integrable losses. Analytic statements use real arithmetic except where numerical errors are explicitly bounded; failed or unreturned evaluations remain failures.

For a probe target, let $y_0,y_+,y_-$ denote the parent and its paid endpoint values, and set $f_b=\min(y_0,y_+,y_-)$. For an already evaluated candidate $j$ with value $y_j$, three improvements are

\[
\Delta_{\rm parent}=y_0-\min(y_0,y_j),\quad
\Delta_{\rm local}=f_b-\min(f_b,y_j),\quad
\Delta_{\rm hist}=h_{\rm pre}-\min(h_{\rm pre},y_j),
\]

where $h_{\rm pre}$ is the best of all paid observations before the candidate is incorporated into the history. A candidate can improve its parent and the local three-point best while leaving the global historical best unchanged.

Let $\mathcal E(s)=1$ denote an eligible state. For a state $s$ with three common observations already paid, let $m_k(s,a,U)$ be the best of $f_b$ and the values obtained by a local action or sequence $a$ using exactly $k$ additional evaluations. The conditional matched-budget contrast is

\begin{equation}\label{eq:local}
L_k(s;a,a')=\mathbb E[m_k(s,a',U_{a'})-m_k(s,a,U_a)\mid s,\ \mathcal E(s)=1].
\end{equation}

Positive values favor $a$. Any random-number coupling must preserve each action's marginal rule. This contrast is conditional on the three common observations already being paid. J2 reports aggregated outcome ratios and normalized decreases, rather than estimates of the raw difference $L_k$.

For an already evaluated point and budget $B$, let $T_K(s)$ commit it and $T_D(s)$ remove only that commitment plus required consistency updates. Let $H_B(T,U;\pi)$ be the best paid value after continuation under policy $\pi$ and a specified continuation stream $U$. The total effect of committing the point is

\begin{equation}\label{eq:commitment}
V_B(s;\pi,\mu_s)=\mathbb E_{U\sim\mu_s}
[H_B(T_D(s),U;\pi)-H_B(T_K(s),U;\pi)].
\end{equation}

A positive value favors KEEP. A single paired replay measures the contrast for one stream; estimating its expectation requires a distribution of streams. Equal seeds define a coupling, but the resulting paths may consume random numbers differently and use different actions or donors.

J6, M3 and CT1 use a separate log-scale quantity, defined when both logarithm arguments are positive:
\begin{equation}\label{eq:logcommitment}
W_B(s;\pi,\mu_s,\epsilon)=
\mathbb E_{U\sim\mu_s}\!\left[
\ln\!\left(\frac{H_B(T_D(s),U;\pi)+\epsilon}
{H_B(T_K(s),U;\pi)+\epsilon}\right)\right].
\end{equation}
A positive $W_B$ favors KEEP on the log-ratio scale. Transformation and averaging are not interchangeable: mean raw differences and mean log ratios estimate different quantities and can have opposite signs even with arbitrarily many futures.

\begin{figure}[!htbp]
\centering
\includegraphics[width=\linewidth]{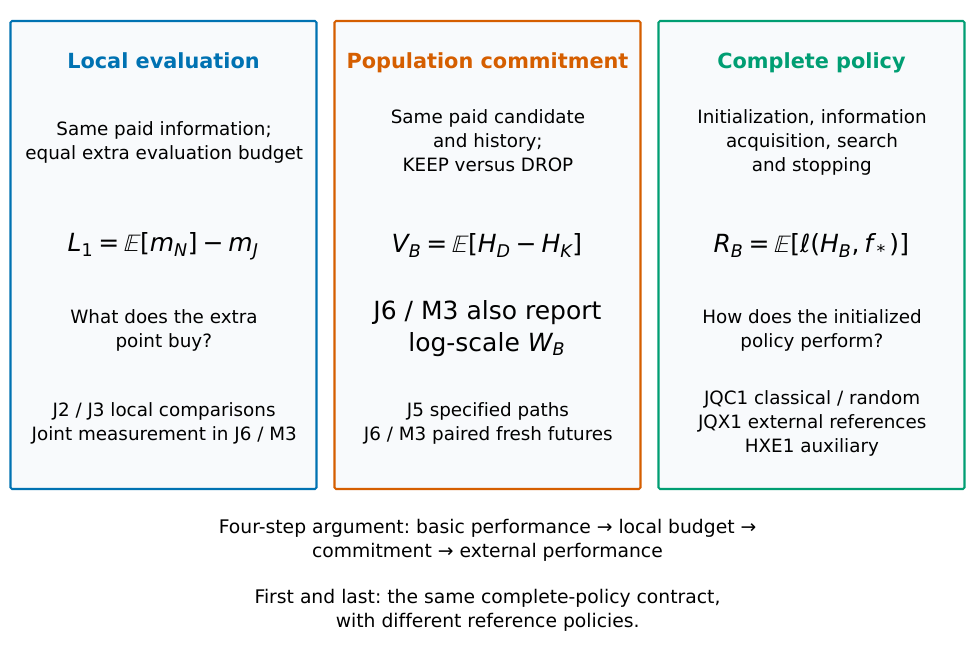}
\caption{Three comparison contracts. Local evaluation value compares equal-cost actions after common observations have been paid. Commitment changes a population slot while preserving paid history. Complete-policy performance includes initialization, information acquisition and stopping. The local and log-continuation estimators are defined in the text; the schematic shows comparison structure, not empirical outcomes.}
\label{fig:estimands}
\end{figure}

For complete policies, define a prespecified loss $\ell$ and a distribution $P$ over problems. Let $P_\pi$ describe each policy's own initialization and random streams conditional on a problem. Full-policy risk is

\begin{equation}\label{eq:policy-risk}
R_B(\pi;P,\ell)=\mathbb E_{f\sim P,\,(x_0,\xi)\sim P_\pi(\cdot\mid f)}
[\ell(H_B(\pi),f_*)],
\end{equation}

where $f_*$ is an offline reference optimum when known. This risk includes initialization, information acquisition, actions and stopping. A cap need not be fully consumed after native early stopping. The conditional local quantity $L_k$, the commitment effects $V_B$ and $W_B$, and $R_B$ differ in state conditioning, treatment, horizon and cost. Figure~\ref{fig:estimands} summarizes the corresponding comparison contracts.

\FloatBarrier
\subsection{Directional interpolation}
For an unreflected direction $d$ and $\phi(t)=f(x+td)$, the quadratic through paid values $y_-=\phi(-1)$, $y_0=\phi(0)$ and $y_+=\phi(1)$ is

\[
q(t)=y_0+\tfrac12(y_+-y_-)t+\tfrac12 ct^2,\quad
c=y_++y_--2y_0,\quad
\alpha=\frac{y_--y_+}{2c}.
\]

Completing the square gives $q(z)-q(\alpha)=c(z-\alpha)^2/2$. Thus the improvement at a node is exactly its positive squared distance from the vertex, scaled by the fitted curvature. When $c>0$ and $0<|\alpha|<1$, the interpolant at $\alpha$ is strictly lower than its values at all three nodes. If $\phi$ is exactly quadratic on the direction, the same conclusion holds for $\phi$. For a non-quadratic $\phi\in C^3[-1,1]$ with $|\phi'''|\le M$,

\[
|\phi(t)-q(t)|\le \frac M6|t(t^2-1)|\le\frac{M}{9\sqrt3}.
\]

Positive fitted curvature alone therefore does not guarantee that the candidate improves the true objective. If the model gap to each node exceeds the corresponding error bound, strict improvement follows; otherwise its evaluated value is required. Sample-value and coordinate errors need additional bounds, given in Appendix A.

The implementation rejects reflected endpoint pairs, non-finite quantities, insufficient positive curvature, fractions too close to zero or endpoints, and mapped candidates coinciding with a node. It checks approximate physical midpoint symmetry after mapping. These guards address eligibility and finite-precision degeneration; they do not bound an arbitrary objective's model error. Reflecting a pair can destroy its symmetry about the original center.

\subsection{Why a local improvement can reverse after commitment}\label{sec:counterexample}
The sweep uses population coordinates to form later trials. Replacing one slot therefore changes more than the retained fitness of that slot. The following exact construction makes that dependency explicit.

Take $f(x)=\|x-(1/2,1/2)\|^2$ on $[0,1]^2$ with four initial slots
\[
x_0=(.68,.80),\quad x_1=(.20,.10),\quad x_2=(.30,.70),\quad x_3=(.28,.50).
\]
Their values are .1224, .25, .08 and .0484. A two-sided probe of slot 0 along $(.1,-.1)$ yields endpoint values .1184 and .1664. Its fitted curvature is .04 and its vertex fraction is .6, giving $j=(.74,.74)$ with value .1152. KEEP writes this interpolant; DROP retains the successful endpoint $(.78,.70)$ with value .1184. Both arms retain the same paid observations, historical best .0484, radii, quota counters and success bookkeeping. A simultaneous probe of slot 2 along $(.1,.1)$ has equal endpoint values .10 and an ineligible center vertex. The policy uses quota allocation, pair-based radius feedback and full eligible interpolation; both probe directions have norm $0.1\sqrt2$. Four initialization calls and five probe calls leave four sweep calls under $B=13$.

The elite slots are $\{2,3\}$ in both arms. Use the valid target/elite/donor tuples $(0,2,1,3)$, $(1,2,0,3)$, $(2,3,0,1)$ and $(3,2,0,1)$, with $F=.5$ and both coordinates crossed. Table~\ref{tab:counterexample} gives the resulting trials and terminal outcomes.
\begin{table}[!htbp]
\centering
\caption{Exact continuation after a locally better commitment. Each displayed trial occurs twice in its arm. All points are feasible and all donor tuples satisfy the distinct-slot restrictions.}
\label{tab:counterexample}
\begin{tabular}{llll}
\toprule
Arm & Improved trials & Rejected trials & Terminal best \\
\midrule
KEEP & $(.48,.52)$, $f=.0008$ & $(.56,.92)$, $f=.18$ & .0008 \\
DROP & $(.50,.50)$, $f=0$ & $(.58,.90)$, $f=.1664$ & 0 \\
\bottomrule
\end{tabular}
\end{table}
The committed candidate is locally better, yet its subsequent historical best is worse. This proves failure of a universal pathwise implication for the population rule. The construction specifies an admissible ideal-real sequence, not the probability of that sequence or the sign of an expected effect. Appendix B gives the full arithmetic.

An ordering can transfer under stronger assumptions. If equal-cost local outcomes obey $m_a\le m_b$ almost surely under a valid coupling, share old best $h_0$, and have the same coupled future candidate minimum $Z$, then
\[
\min(h_0,m_a,Z)\le\min(h_0,m_b,Z).
\]
Integrable expectations preserve this pointwise order. Population-dependent futures generally violate the common-$Z$ condition, as the construction shows. For KEEP/DROP with identical paid history, a truly common future sequence instead yields zero commitment effect. Appendix A also gives an exact example showing why an ordering of local means alone is insufficient.

\subsection{When a commitment effect determines policy performance}\label{sec:risk-identity}
Consider two complete policies $\pi_K$ and $\pi_D$ under the same problem distribution. Assume they have identical initialization and prefix policy, yielding the same distribution $Q$ of a fully specified pre-intervention state $S$. They differ only in one designated KEEP/DROP treatment of an already paid candidate, then follow the same continuation policy, conditional random-stream law and logical FE contract. Represent a trajectory that never reaches the designated treatment by a null state, on which the two policies coincide. For the prespecified integrable loss, define
\[
V_B^{\ell}(s)=\mathbb E\!\left[
\ell(H_B(T_D(s),U;\pi),f_*)-
\ell(H_B(T_K(s),U;\pi),f_*)\mid S=s\right],
\]
with $V_B^{\ell}=0$ on the null state. The law of total expectation gives
\[
R_B(\pi_D;P,\ell)-R_B(\pi_K;P,\ell)
=\int V_B^{\ell}(s)\,Q(ds).
\]
Indeed, condition each complete-policy loss on the common prefix state, subtract the two conditional expectations, and integrate over their common state law. For $\ell(h,f_*)=h-f_*$ this is the mean of the raw commitment contrast $V_B$; other losses require the corresponding $V_B^{\ell}$. If eligibility has probability $p>0$ and null trajectories have zero contrast, the integral equals $p\,\mathbb E[V_B^{\ell}(S)\mid\text{eligible}]$, not the selected-state mean alone. If $p=0$, the policies coincide on $Q$-almost every trajectory and their risk difference is zero; the conditional expectation given eligibility need not be defined.

These common-prefix, single-treatment assumptions determine when commitment effects can be averaged into a policy comparison. NQ and JQ100 also differ in information acquisition and subsequent states, while external solvers differ more broadly. Their risk differences require complete-policy comparisons. The loss must also match: a raw-value effect cannot replace a log-loss effect.

\section{Experimental design}
Table~\ref{tab:question-map} maps the four questions to their sampling units. JQ100 is fixed across the original local studies and complete-policy comparisons; follow-up studies change the sampling or comparison design.

\begin{table}[!htbp]
\centering\small
\caption{Questions, comparisons and sampling units in the four-step argument. Study labels distinguish experiments using different sampling and comparison designs.}
\label{tab:question-map}
\begin{tabular}{>{\raggedright\arraybackslash}p{.18\linewidth}>{\raggedright\arraybackslash}p{.36\linewidth}>{\raggedright\arraybackslash}p{.35\linewidth}}
\toprule
Step & Comparison and study & Unit and principal scope \\
\midrule
Basic effectiveness & Classical references (JQC1); same start (PC1) & Complete run; eight profiles, 10D/20D, $1000D$ FE \\
Local budget value & One/two-point branches (J2); 24-FE branches (J3) & Eligible state with three common observations already paid \\
Population commitment & Accepted-state DGO/DE replay (J6/M3); preselected types (CT1) & State, with four futures (J6/M3) or two (CT1); different populations \\
Complete performance & Same-policy references (JQX1); public suite (BX1); restart (PC1) & Complete run; native stopping or separately specified restart \\
\bottomrule
\end{tabular}
\end{table}

\subsection{Complete policies, benchmarks and initialization}
Table~\ref{tab:complete-designs} summarizes the complete-policy studies. JQC1 compares classical references; PC1 tests matched initialization, restart and parameter changes; JQX1 compares JQ100 with internal and strong references; BX1 tests public-suite transfer. All use dimensions 10 and 20 and a cap of $1000D$ FE.

\begin{table}[!htbp]
\centering\small
\caption{Complete-policy designs. Instances and search seeds are counted per profile or function and dimension. Run totals include both dimensions and every listed policy. PC1's four perturbations change one JQ100 parameter at a time.}
\label{tab:complete-designs}
\begin{tabular}{l>{\raggedright\arraybackslash}p{.21\linewidth}c r >{\raggedright\arraybackslash}p{.35\linewidth}}
\toprule
Study & Problems & Instances $\times$ seeds & Runs & Policies \\
\midrule
JQC1 & 8 profiles, 6 families & $3\times5$ & 960 & JQ100, DE5D, PSO, Random \\
PC1 & Same 8 profiles & $3\times3$ & 1,296 & JQ100, DE40, DE5D, CMA-native/restart, 4 perturbations \\
JQX1 & 5 custom conditions & $3\times3$ & 450 & JQ100, NQ, S0, L-SHADE, CMA-ES \\
BX1 & 24 boxed BBOB functions & $3\times2$ & 1,728 & JQ100, DE5D, PSO, Random, L-SHADE, CMA-ES \\
\bottomrule
\end{tabular}
\end{table}

JQ100, PSO and Random share 40 initial points in JQC1 and BX1; DE5D uses all $5D$ points of the same master array. PC1 adds DE40 with exactly JQ100's 40-point start. JQX1 shares 40 starting points among JQ100, NQ and S0. L-SHADE and CMA-ES retain native initialization and stopping.

The custom profiles cover sphere, ellipsoid, Rosenbrock, Rastrigin, Ackley and doublewell landscapes, with axis-aligned and rotated variants where specified. They test directional fit, ill-conditioning, curved valleys and basin trapping. Fresh shifts and rotations give new instances within these known families. BX1 uses all 24 official COCO \texttt{bbob-boxed} functions \citep{cocoboxed}, including nonzero objective offsets. Solvers receive raw objective values; known optima are used only in retrospective error analysis. Appendix F lists the custom profiles and targets.

Random tests progress beyond uniform sampling; DE/rand/1/bin and PSO are classical population references; L-SHADE and CMA-ES test stronger competitiveness. JQ100 underwent earlier development-stage selection, while the classical parameters were fixed. DE uses $F=.5$, crossover .9, reflection and strict replacement; PC1 changes its population size to 40 while retaining these rules. Appendices F and~\ref{app:followup} specify the remaining baseline settings. The comparison is between these configured policies, without equal tuning histories.

\subsection{Local branches and state-consistent continuations}
J2 compares one- and two-point local actions after evaluating the parent and endpoints; J3 compares three 24-FE local policies on separate states. For continuation studies, a \emph{base run} generates the state to be intervened on. J6 selects the first accepted eligible interpolation in a prescribed budget window from each of 48 DGO runs. M3 adds one diagnostic probe to DE/rand/1/bin and selects states from 29 of 32 runs. Each selected state has four paired future streams and 16 fresh-direction local references. Replaying KEEP with the original stream must recover the original path before new continuations are used. Runs without a selected state remain in the sampling denominator.

At the intervention, KEEP and DROP share the evaluated third point, history, historical best, FE count, radius feedback, scores and random-generator state. DROP restores the slot that would have remained without the third-point commitment and updates dependent state fields. Subsequent differences include all effects through donors, radii and allocation.

CT1 preselects 128 NQ states from four profiles: two rotated ellipsoids, rotated Rosenbrock and rotated Rastrigin, with 16 instances per profile and dimension. The slot and direction are chosen before checking eligibility. At each state, it evaluates two endpoints and four candidate types: interpolation, random interior, new direction and the better endpoint. All six evaluations enter a shared history, including repeated points and the parent fallback for ineligible interpolation.

For each candidate type, KEEP writes a strict parent improvement and DROP retains the parent; two paired futures continue both states to a $500D$ cap. Nonaccepting KEEP states coincide with DROP and remain in the analysis. CT1 therefore estimates type-specific writeback after a common paid diagnostic batch, rather than a policy that generates only its own candidate. Its parent-improvement rule differs from JQ100's requirement to beat both parent and better endpoint. A second interpolation pair holds all checkpoint radii fixed to test interaction with radius adaptation. J6/M3 instead select accepted states and use $1000D$ budgets.

\subsection{Metrics, uncertainty and scope of inference}
Endpoint ratios are geometric means of paired error ratios under the stated problem weights. JQC1 and PC1 truncate errors at each profile's target and weight six families equally. JQX1 uses a $10^{-12}$ floor. BX1 uses $\max(e,0.01)$ and weights all 24 functions equally. Ratios below one favor the numerator. Target truncation treats errors below the required tolerance alike; a numerical floor retains finer precision down to that floor.

The continuation log ratios use nonnegative, zero-optimum custom objectives with an additive $10^{-12}$ constant. They depend on the objective origin and near-zero convention. Public-suite errors instead subtract the known optimum. We report target attainment, endpoint ratios and adverse tails together; S1 examines sensitivity to their numerical representation using saved data.

Uncertainty estimates preserve the study's pairing and sampling structure. J6/M3 and CT1 sample states, with paired futures nested within each state. Retrospective PC1 and BX1 analyses resample shared search-seed blocks across all profiles or functions, and instances within them. Their three and two observed seed blocks define the empirical variation covered. Functions remain fixed, dimensions are reported separately, and intervals are pointwise. Prespecified improvement and tail thresholds are practical screening rules, not significance or equivalence tests. Appendix~\ref{app:followup-statistics} gives the full resampling schemes.

\section{Results}
\subsection{Basic effectiveness and matched-start diagnostics}
JQ100 made useful progress within the allotted budget, meeting the prespecified basic screen against Random and improvement screens against fixed DE and PSO in both dimensions (Table~\ref{tab:jqc1}). Aggregate gains coexisted with losses on individual profiles and runs: 10D axis-aligned Rastrigin favored PSO, and a 20D doublewell pair had a target-truncated JQ100/PSO ratio of $2\times10^5$.

\begin{table}[!htbp]
\centering
\caption{JQ100 versus fixed references in JQC1. Ratios below one favor JQ100; intervals are descriptive 95\% hierarchical bootstrap intervals. Hits are raw counts out of 120 per method. Basic and classical screens use family-weighted hits and their respective prespecified rules. In JQ100/DE/PSO/Random order, family-weighted hit rates (\%) are 59.44/41.67/47.22/0 in 10D and 56.11/0/38.33/0 in 20D.}
\label{tab:jqc1}
\begin{tabular}{llrll l}
\toprule
$D$ & Reference & Ratio & 95\% interval & Hits JQ/ref & Criteria \\
\midrule
10 & Random & $1.006\times10^{-5}$ & $[9.299\times10^{-6},1.082\times10^{-5}]$ & 62/0 & Basic met \\
10 & DE & 0.3727 & [0.3276, 0.4211] & 62/45 & Improvement met \\
10 & PSO & 0.1256 & [0.0602, 0.2590] & 62/50 & Improvement met \\
20 & Random & $9.449\times10^{-6}$ & $[6.979\times10^{-6},1.480\times10^{-5}]$ & 58/0 & Basic met \\
20 & DE & 0.00219 & [0.00157, 0.00352] & 58/0 & Improvement met \\
20 & PSO & 0.1506 & [0.0518, 0.4719] & 58/42 & Improvement met \\
\bottomrule
\end{tabular}
\end{table}

Matching DE's initial population to JQ100's 40 points narrowed the mean advantage. The matched-start intervals span parity in both dimensions, and the family results include an adverse 10D doublewell comparison (Tables~\ref{tab:pc1-main}--\ref{tab:pc1-families}). DE40 also outperformed DE5D in aggregate. Population size changes both initialization and later dynamics, so this difference cannot be assigned to the starting points alone.

\begin{table}[!htbp]
\centering\small
\caption{Same-start diagnosis in PC1. Ratios are six-family target-truncated geometric means under the same $1000D$ cap. Brackets show retrospective 95\% percentile intervals preserving shared search-seed blocks; three observed seeds define their empirical scope. The final column gives the worst paired target-truncated ratios at 10D/20D.}
\label{tab:pc1-main}
\begin{tabular}{llll}
\toprule
Comparison & 10D [interval] & 20D [interval] & Worst 10D/20D \\
\midrule
JQ100/DE40 & 0.6154 [0.4476, 1.23] & 0.4446 [0.1172, 1.711] & $2.00\times10^5$ / $2.00\times10^5$ \\
JQ100/DE5D & 0.2727 [0.1243, 0.4049] & 0.002412 [0.001373, 0.006559] & 1.077 / 5.528 \\
DE40/DE5D & 0.4431 [0.1666, 0.8073] & 0.005426 [0.002241, 0.01392] & 1.942 / 5.819 \\
\bottomrule
\end{tabular}
\end{table}
\begin{table}[!htbp]
\centering\small
\caption{All PC1 families for the matched-start JQ100/DE40 comparison. Ratios use target-truncated errors; hits are JQ100/DE40 counts. Ellipsoid and Rastrigin each contain two profiles and 18 runs per dimension; other families contain nine. The six families have equal macro weight.}
\label{tab:pc1-families}
\begin{tabular}{lrrrr}
\toprule
Family & 10D ratio & 10D hits & 20D ratio & 20D hits \\
\midrule
Sphere & 1 & 9/9 & 1 & 9/9 \\
Ellipsoid & 0.08621 & 11/9 & 0.5549 & 9/9 \\
Rosenbrock & 0.6816 & 0/0 & 0.8739 & 0/0 \\
Rastrigin & 0.2383 & 0/0 & 0.2399 & 0/0 \\
Ackley & 1 & 9/9 & 1 & 9/9 \\
Doublewell & 3.882 & 8/9 & 0.06637 & 7/5 \\
\bottomrule
\end{tabular}
\end{table}

DGO is therefore a useful test optimizer for the mechanism questions. The next steps ask which evaluations help locally and whether committing their candidates improves later search.

\subsection{Local value under matched expenditure}
At the same eligible state, spending the next evaluation on interpolation outperformed a new-direction point on the quadratic group. J2 met its conditional local criteria at both one- and two-evaluation budgets (Table~\ref{tab:local-evidence}). The curved group did not meet the joint criterion, and multimodal results varied by condition.

J3 tested whether the interpolation model helped over a longer local horizon. At 24 FE, neither fresh-pair/model reuse versus interpolant-first nor reuse versus no reuse met the joint criterion in either direction. A favorable four-point prefix did not persist as support at the full horizon. Both studies leave the surrounding population outside the local experiment; they motivate a separate test of what happens when the candidate is committed.

\begin{table}[!htbp]
\centering
\caption{Quadratic-group matched local comparisons. Entries are numerator/denominator outcome ratios, with smaller values favoring the first policy. NQ and JQ identify the base-run populations. The studies use separate selected states and their own aggregation rules.}
\label{tab:local-evidence}
\begin{tabular}{llll}
\toprule
Study and extra FE & Comparison & NQ base run & JQ base run \\
\midrule
J2, 1 FE & Interpolation / new direction & 0.951 & 0.944 \\
J2, 2 FE & Interpolation + direction / two directions & 0.956 & 0.945 \\
J3, 24 FE & Fresh-pair reuse / interpolant first & 1.019 & 1.015 \\
J3, 24 FE & Fresh-pair reuse / no reuse & 0.996 & 0.980 \\
\bottomrule
\end{tabular}
\end{table}
\FloatBarrier

\subsection{Population commitment}
Earlier J1 tests showed aggregate quadratic gains alongside a severe individual loss. Two historical KEEP/DROP replays then produced opposite continuation outcomes (Appendix G). These cases motivated new-state experiments with multiple paired futures.

\begin{figure}[!htbp]
\centering
\includegraphics[width=\linewidth]{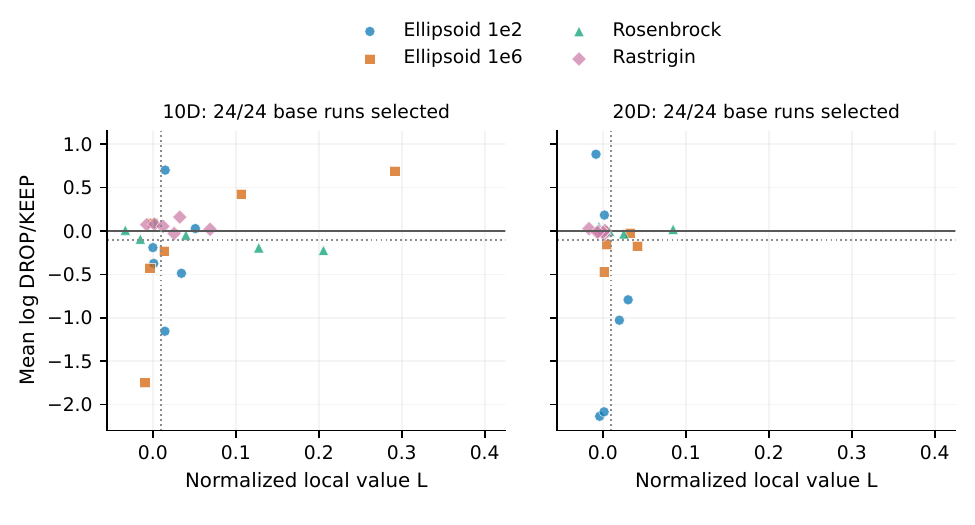}
\caption{J6 local and continuation estimates: 48 selected states, one point per base run and 24 per dimension. Horizontal: normalized mean one-point advantage over 16 fresh-direction references. Vertical: mean $\ln[(h_D+10^{-12})/(h_K+10^{-12})]$ over four paired futures. Dotted lines mark $0.01$ and $\ln(0.9)$; axes match Figure~\ref{fig:m3}. Markers may overlap.}
\label{fig:j6}
\end{figure}

\begin{figure}[!htbp]
\centering
\includegraphics[width=\linewidth]{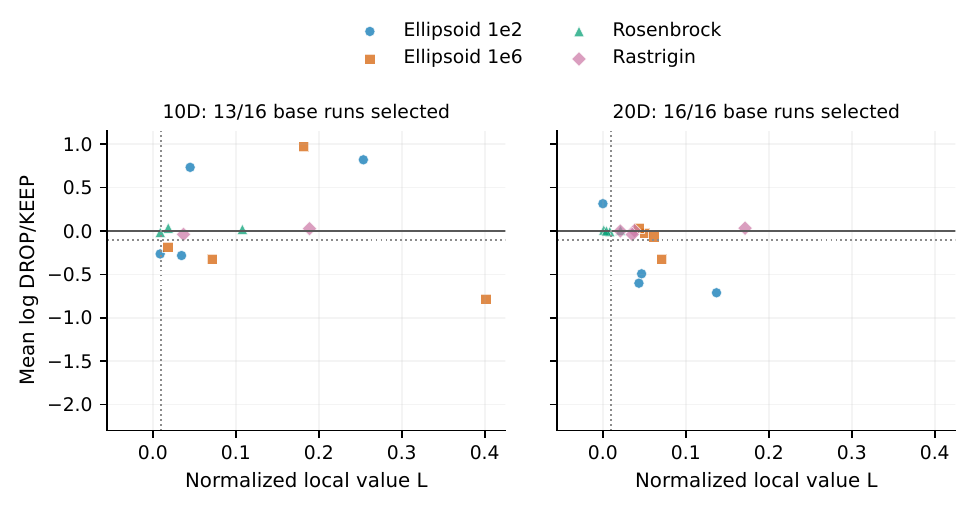}
\caption{M3 local and continuation estimates: 13/16 base runs selected at 10D and 16/16 at 20D. Unselected runs are omitted, not assigned zero effects. Coordinates, references, four-future averaging, thresholds and axis limits follow Figure~\ref{fig:j6}; markers may overlap.}
\label{fig:m3}
\end{figure}

\paragraph{Accepted-state replays on DGO and DE.}
J6 and M3 test whether the distinction between local gain and continuation depends on DGO's scheduling rule. Figures~\ref{fig:j6}--\ref{fig:m3} show locally favorable candidates with both favorable and adverse mean continuations in the DGO and DE-plus-probe settings. The primary quadratic-group intervals span zero (Table~\ref{tab:commit-main}). Four paired futures per state reveal variation in estimated outcomes but do not determine each state's true expected sign.

\begin{table}[!htbp]
\centering
\caption{Primary quadratic commitment estimates on accepted states. Positive mean log DROP/KEEP favors KEEP. Each state has four paired futures; these repeats do not multiply the number of sampled states.}
\label{tab:commit-main}
\begin{tabular}{llcl}
\toprule
Study & Dimension & States/base runs & Mean log ratio [95\% interval] \\
\midrule
DGO replay (J6) & 10 & 12/12 & $-0.223\;[-0.764,0.279]$ \\
DGO replay (J6) & 20 & 12/12 & $-0.484\;[-1.131,0.153]$ \\
DE transfer (M3) & 10 & 8/8 & $0.085\;[-0.464,0.647]$ \\
DE transfer (M3) & 20 & 8/8 & $-0.235\;[-0.537,0.049]$ \\
\bottomrule
\end{tabular}
\end{table}

\FloatBarrier

\begin{figure}[!htbp]
\centering
\includegraphics[width=\linewidth]{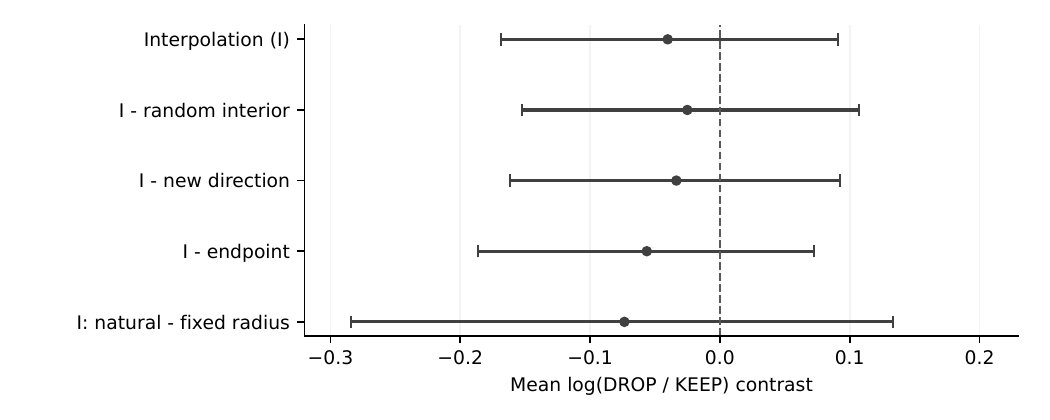}
\caption{CT1 quadratic-group commitment effects and paired contrasts across 64 preselected states. Bars show 95\% pointwise hierarchical bootstrap intervals. Positive log DROP/KEEP favors KEEP for interpolation I; positive contrasts indicate a larger log DROP/KEEP effect for the first candidate type or radius policy. All states, including no-action outcomes, enter the estimates; intervals spanning zero do not establish equivalence.}
\label{fig:followup-summary}
\end{figure}

\paragraph{Preselected states, candidate types and radius policy.}
CT1 broadens the comparison to preselected states and alternative candidate types. Its interpolation effect, type contrasts and radius-policy contrast all have quadratic-group intervals spanning zero (Figure~\ref{fig:followup-summary}). The interpolation interval met the prespecified width target but still includes practically relevant benefit and harm. Acceptance varies markedly by type, with little overlap across all four types (Table~\ref{tab:ct1-acceptance}); the all-state contrasts therefore combine eligibility, acceptance and subsequent writeback effects.

Every accepted interpolant also improved the better original endpoint. Adding that requirement leaves the accepted set and all paired outcomes unchanged in this sample (Table~\ref{tab:ct1-acceptance}). The comparison still concerns CT1's shared diagnostic history and parent-retaining DROP rule.

\begin{table}[!htbp]
\centering\small
\caption{CT1 acceptance among preselected states. I, R, N and E denote interpolation, random interior, new direction and the better endpoint. ``Strict I'' additionally requires improvement over the better original endpoint. ``All four'' counts states accepting every type.}
\label{tab:ct1-acceptance}
\begin{tabular}{lrrrrrrr}
\toprule
Group & States & I & Strict I & R & N & E & All four \\
\midrule
Quadratic & 64 & 57 & 57 & 13 & 8 & 14 & 1 \\
Controls & 64 & 43 & 43 & 20 & 13 & 20 & 1 \\
All states & 128 & 100 & 100 & 33 & 21 & 34 & 2 \\
\bottomrule
\end{tabular}
\end{table}

\FloatBarrier
\subsection{Complete solver performance}
A complete solver must reach useful states, allocate evaluations and stop. We now compare initialized policies, first on the mechanism-oriented custom conditions and then on the public boxed suite.

\begin{figure}[!htbp]
\centering
\includegraphics[width=\linewidth]{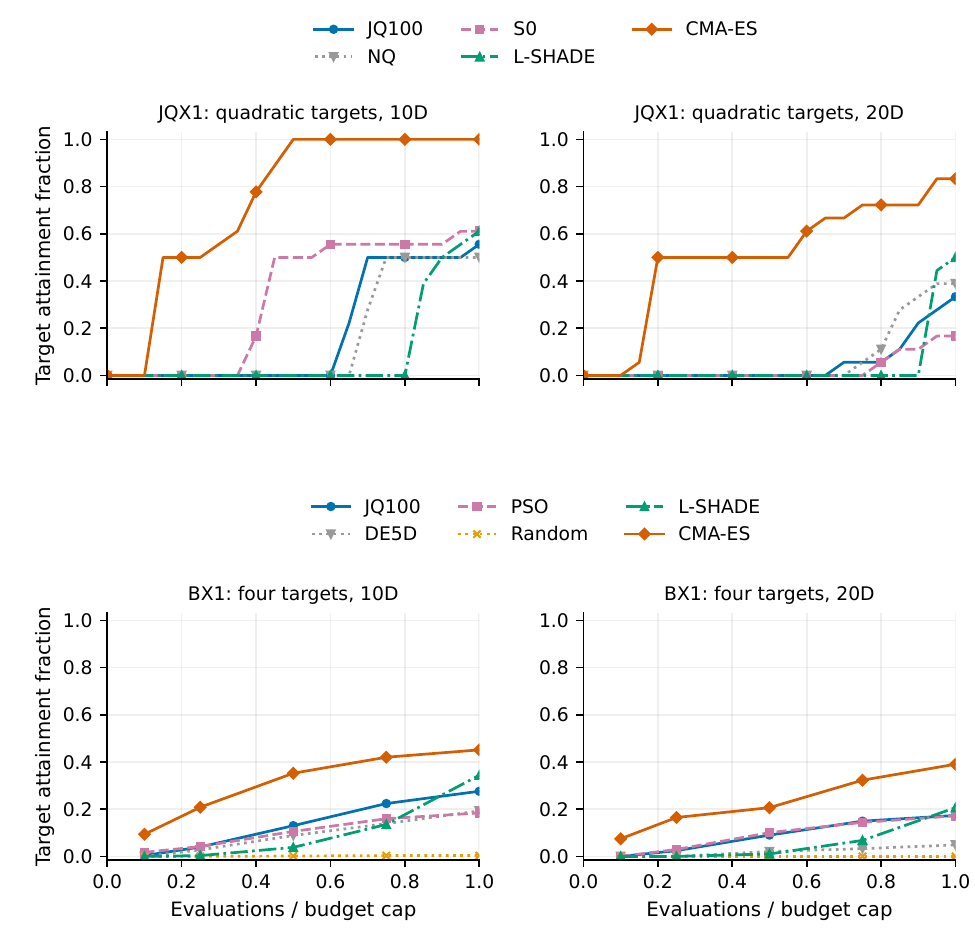}
\caption{Target attainment by evaluation budget. Top: JQX1 quadratic targets, 18 runs per method and dimension; first hits use batch-end upper bounds. Bottom: BX1, 144 runs per method and dimension, averaged over targets $1,10^{-2},10^{-4},10^{-8}$ (576 run--target pairs); hits use recorded evaluation order. Lines connect the budget grid. Native stopping adds neither evaluations nor later hits.}
\label{fig:target-arrival}
\end{figure}

\paragraph{Same-version strong references and target arrival.}
JQ100 did not meet the external comparison criteria against L-SHADE or CMA-ES in either dimension. Its internal advantage over NQ also reversed at 20D (Table~\ref{tab:jqx1}). CMA-ES reached quadratic targets earlier and more often, despite frequent native stopping (Figure~\ref{fig:target-arrival}, top). Some doublewell and Rastrigin comparisons favored JQ100 (Figure~\ref{fig:jqx1}), but the aggregate stress criterion was not met. Large floor-based ratios in Table~\ref{tab:jqx1} reflect very fine terminal precision; target attainment gives their practical context. Appendix C reports the separate H10 configuration.

\paragraph{Public-suite transfer.}
The public boxed suite gives the same aggregate ordering: JQ100 has lower target-truncated errors than the tested DE, PSO and Random policies, and higher errors than L-SHADE and CMA-ES (Table~\ref{tab:bx1}). Shared-seed resampling and separate seed summaries retain that direction. Individual losses remain, including severe tails against PSO (Table~\ref{tab:bx1-tails}). Multi-target counts and curves show how the comparison changes with required precision (Table~\ref{tab:bx1-multitarget}; Figure~\ref{fig:target-arrival}, bottom). Corrected initial records leave the terminal results unchanged (Appendix~\ref{app:followup-integrity}).

\begin{table}[!htbp]
\centering\small
\caption{BX1 JQ100/reference geometric error ratios on 24 official boxed BBOB functions, using $\max(e,0.01)$. Each dimension has three instances and two shared search seeds per function. Brackets are retrospective 95\% percentile intervals preserving the shared seed blocks across functions. They describe sensitivity within this two-seed design.}
\label{tab:bx1}
\begin{tabular}{lll}
\toprule
Reference & 10D ratio [interval] & 20D ratio [interval] \\
\midrule
DE5D & 0.3667 [0.278, 0.4742] & 0.1349 [0.1139, 0.1695] \\
PSO & 0.2464 [0.1868, 0.302] & 0.5877 [0.4989, 0.6965] \\
Random & 0.001489 [0.001135, 0.001866] & 0.001876 [0.00158, 0.002358] \\
L-SHADE & 1.653 [1.235, 2.204] & 1.897 [1.529, 2.531] \\
CMA-ES & 1.681 [1.169, 2.306] & 4.706 [3.401, 6.421] \\
\bottomrule
\end{tabular}
\end{table}

\begin{table}[!htbp]
\centering\small
\caption{BX1 terminal target attainment from the same saved runs. Each entry lists hit counts at errors $1,10^{-2},10^{-4},10^{-8}$, in that order, out of 144 runs per method and dimension. These fixed tolerances complement the primary $10^{-2}$-truncated endpoint ratio.}
\label{tab:bx1-multitarget}
\begin{tabular}{lll}
\toprule
Method & 10D hits & 20D hits \\
\midrule
JQ100 & 81 / 41 / 25 / 12 & 56 / 26 / 12 / 6 \\
DE5D & 61 / 28 / 19 / 3 & 23 / 5 / 0 / 0 \\
PSO & 54 / 27 / 18 / 7 & 41 / 27 / 19 / 12 \\
Random & 2 / 0 / 0 / 0 & 0 / 0 / 0 / 0 \\
L-SHADE & 86 / 54 / 38 / 20 & 67 / 26 / 18 / 8 \\
CMA-ES & 90 / 64 / 56 / 50 & 82 / 52 / 48 / 43 \\
\bottomrule
\end{tabular}
\end{table}

\paragraph{Stopping, precision and parameter contracts.}
Stopping rules affect both error and cost. Restarting CMA-ES improved terminal errors in PC1 by using more evaluations. In BX1, JQ100 remained worse than CMA-ES when both saved paths were read at the earlier stopping count of each pair (Table~\ref{tab:stopping}). Those pair-specific horizons depend on the observed stopping outcomes, so they describe matched realized costs rather than a new fixed-budget experiment.

\begin{table}[!htbp]
\centering\small
\caption{Stopping-policy comparisons. Ratios use the study-specific target-truncated errors and weights. PC1 compares complete restart/native policies; BX1 reads paired saved paths at the earlier actual stopping count. Native CMA-ES used 2,591,274 FE in BX1, versus 4,320,000 for each other policy.}
\label{tab:stopping}
\begin{tabular}{>{\raggedright\arraybackslash}p{.30\linewidth}rr>{\raggedright\arraybackslash}p{.41\linewidth}}
\toprule
Comparison & 10D ratio & 20D ratio & Evaluation scope \\
\midrule
PC1: CMA-restart/native & 0.370 & 0.317 & 2,160,000 / 994,172 total FE across both dimensions \\
BX1: JQ100/CMA-ES at matched counts & 5.893 & 10.648 & Pair-specific counts: 860--10,000 FE at 10D; 2,640--20,000 at 20D \\
\bottomrule
\end{tabular}
\end{table}

The error representation also matters. In the saved-data analysis, the 10D JQ100/S0 ordering reverses between numerical-floor precision and task-target truncation, while the JQ100/CMA-ES ordering remains unchanged (Table~\ref{tab:s1-scales}). Thus a large floor-based ratio need not represent an equally large practical loss at the required tolerance.

PC1 shows parameter dependence: radius-feedback changes reverse their mean effect between dimensions, while both tested initial-radius changes improve macro means but retain adverse families and tails (Table~\ref{tab:pc1-details}). The main JQ100 configuration remains fixed. A new choice would require independent evaluation.

\section{Discussion}
\subsection{What connects the four steps}
The experiments separate three decisions that a terminal score combines. A useful evaluation can improve the local information available to search; commitment determines which coordinates later population operations use; and a complete policy must acquire those states and allocate its entire budget. DGO provides a working test case in which local gains coexist with uncertain commitment effects and strong-solver deficits.

The exact construction explains how the separation can occur: changing one slot changes later donor-based trials. The risk identity gives the bridge to policy performance: with a common prefix, one treatment and a matched loss, the statewise effects integrate to the policy-risk difference. Repeated interpolation and different external solvers change the state distribution and information costs as well, so their performance must be measured directly.

\subsection{What the intervention identifies}
The KEEP/DROP intervention estimates the total continuation effect of population commitment while holding the paid information fixed. It includes all later changes in donors, radii and allocation. The observed variation does not identify a single pathway: CT1's type and radius-policy contrasts remain uncertain, and few states accept every candidate type. This distinguishes a test of commitment from a mechanism-specific attribution.

A practical commitment rule would need predictors available when the candidate is evaluated. Its full-policy test would then include prediction costs, missed opportunities and accumulated state changes. Outcomes observed only after replay can diagnose a path but cannot guide that earlier decision.

\subsection{Scope and interpretation}
The empirical scope is deterministic 10D and 20D optimization on the stated custom families and fixed public suite. Results depend on target precision, initialization, stopping and parameters. The matched-start and restart studies measure those policy comparisons without decomposing them into isolated causes. Future parameter choices require samples independent of those used for selection.

The mathematical results establish a valid finite-budget process, sufficient interpolation conditions, a possible ordering reversal and a conditional link to policy risk. Average empirical effects are governed by their own sampling designs and intervals. Together, these provide complementary forms of evidence for the three comparison questions.

\section{Conclusion}
DGO demonstrates why local budget gains, population commitment and solver performance need separate evaluation. Short-horizon matched comparisons favor interpolation on the tested eligible quadratic states, whereas the separate 24-FE local study does not establish a sustained advantage. Continuation estimates remain mixed, and aggregate gains over fixed classical references do not extend to the strong references. The matched-start intervals also span parity. Defining the state, cost, loss and horizon makes these outcomes interpretable together.

\paragraph{Code availability and third-party components.}
The code, analysis scripts and supporting results will be made available at \url{https://github.com/Richardlsq/dgo-mechanism-study}. Project code is intended for release under the MIT license. Third-party software retains its own license terms. The L-SHADE author distribution has no explicit redistribution license; its source will not be redistributed, and instructions for obtaining the original implementation will accompany the comparison code.

\paragraph{Funding.}
This research received no specific grant from any funding agency in the public, commercial or not-for-profit sectors.

\paragraph{Competing interests.}
The author declares no competing interests.

\paragraph{Use of generative AI and author responsibility.}
OpenAI Codex assisted with code implementation, experimental orchestration, verification and statistical analysis, source organization, mathematical formalization, symbolic checks and manuscript drafting. The human author is responsible for the scientific content and the accuracy of this disclosure.

\appendix
\section{Sufficient conditions and model error}

\subsection*{Interpolation conditions and error propagation}
Let $\phi(t)=f(x+td)$ on $[-1,1]$, with three unreflected ideal nodes $-1,0,1$. Here $x$, $d$, and the coordinate-error bound $\epsilon_x$ use physical coordinates and the Euclidean norm; $G$ is the Lipschitz constant in that same norm. In the implementation $d=\mathrm{span}\odot\delta$ for unit-box direction $\delta$. A unit-box error bound must first be multiplied by the Euclidean operator norm of $\operatorname{diag}(\mathrm{span})$. The bound $M$ below is on the third derivative with respect to the parameter $t$. For the exact-data quadratic interpolant, completing the square gives
\[
q(z)-q(\alpha)=\frac{c}{2}(z-\alpha)^2.
\]
Consequently $c>0$ and $0<|\alpha|<1$ give strict improvement at all three nodes for $q$, and for $\phi$ when $\phi=q$. For general $\phi\in C^2[-1,1]$, integration by parts on $[-1,0]$ and $[0,1]$ gives
\[
c=\int_{-1}^{1}(1-|u|)\phi''(u)\,du.
\]
Thus a positive fitted curvature is not a certificate of pointwise positive curvature along the entire segment.

If $\phi\in C^3[-1,1]$ and $|\phi'''|\le M$, the interpolation remainder is bounded by $M|t(t^2-1)|/6$. Maximizing $|t(1-t^2)|$ on this interval gives $2/(3\sqrt3)$, so write $\varepsilon=M/(9\sqrt3)$. If $c>0$ and $\alpha\in(-1,1)$, a minimizer $t_*$ of $\phi$ on the compact interval exists and
\[
\phi(\alpha)-\phi(t_*)
\le q(\alpha)-q(t_*)+2\varepsilon\le2\varepsilon.
\]
If additionally $\phi''\ge\mu>0$, constrained first-order optimality at $t_*$ and strong convexity imply
\[
\frac{\mu}{2}|\alpha-t_*|^2
\le\phi(\alpha)-\phi(t_*)\le2\varepsilon,
\qquad |\alpha-t_*|\le2\sqrt{\varepsilon/\mu}.
\]
These bounds require $M$ and $\mu$, which the numerical guards do not supply.

Under the same $C^3$ assumption and third-derivative bound, for perturbed values $|y_i-\phi(i)|\le\epsilon_y$, let $q_y$ denote their exact quadratic interpolant, with $c_y>0$ and vertex $\alpha_y$. For the parameter $t\in[-1,1]$ actually used, the absolute sum of its three Lagrange weights is $1+|t|-t^2\le5/4$. Hence $|q_y(t)-q(t)|\le(1+|t|-t^2)\epsilon_y$, whereas $|q_y(z)-\phi(z)|\le\epsilon_y$ at a node $z$. If the actual candidate differs from $x+td$ by at most $\epsilon_x$, and $f$ is $G$-Lipschitz on a neighborhood containing both points, its true value minus $\phi(z)$ is at most
\[
-\frac{c_y}{2}[(z-\alpha_y)^2-(t-\alpha_y)^2]
+\frac M6|t(1-t^2)|
+(2+|t|-t^2)\epsilon_y+G\epsilon_x.
\]
The candidate strictly improves all three true node values if this bound is negative for each node; the numerical curvature threshold does not bound interpolation, sample or coordinate error.

\subsection*{When local ordering transfers}
Suppose equal-cost local alternatives have the same old historical best $h_0$, satisfy $m_a\le m_b$ almost surely under a coupling with valid marginals, and have the same remaining evaluation allowance. If the two future sequences have the same marginal law and are coupled to the same sequence, whose minimum is $Z$ (set $Z=+\infty$ if no evaluations remain), then pointwise
\[
H_a=\min(h_0,m_a,Z)\le\min(h_0,m_b,Z)=H_b.
\]
Integrable raw expectations preserve this ordering. Independence of the future sequence from the local result and state is sufficient for the common-sequence condition, whereas population-dependent candidates need not satisfy it. In KEEP/DROP the paid histories already coincide, so a common future sequence gives identical terminal historical best and zero commitment effect.

An ordering of local means alone is insufficient. Let $h_0=2$, $m_a=1/2$, and let $m_b$ equal $0$ with probability $3/5$ and $2$ with probability $2/5$. With deterministic common future minimum $Z=2/5$,
\[
\mathbb E[m_a]=\tfrac12<\tfrac45=\mathbb E[m_b],
\qquad
\mathbb E[H_a]=\tfrac25>\tfrac4{25}=\mathbb E[H_b].
\]
This arithmetic example concerns expectations under truncation by a minimum.

The single-treatment policy-risk identity and proof are in Section~\ref{sec:risk-identity}.

\section{An exact population-rule counterexample}

Consider $f(x_1,x_2)=(x_1-\tfrac12)^2+(x_2-\tfrac12)^2$ on $[0,1]^2$, with $N=4$, quota allocation, pair-based radius feedback, full interpolation fraction and total budget $B=13$. Start at

\[
x_0=(.68,.80),\quad x_1=(.20,.10),\quad x_2=(.30,.70),\quad x_3=(.28,.50).
\]

The initial values are .1224, .25, .08 and .0484. After four initial evaluations, choose probe targets 0 and 2 with perturbations $(.1,-.1)$ and $(.1,.1)$, each of norm $0.1\sqrt2$. Target 0's endpoints have values .1184 and .1664; its interpolant has $c=.04$, $\alpha=.6$, and candidate $j=(.74,.74)$ with value .1152. Target 2's endpoint values are both .10 and its fitted vertex is the center, so it is ineligible under the interior-fraction guard. The probe costs four endpoint evaluations and one third-point evaluation.

KEEP writes $j$ to slot 0; DROP writes the successful endpoint $(.78,.70)$ with value .1184. Both retain historical best .0484, the same radius, evaluation count and success bookkeeping. Probe has spent five FE and sweep none, so quota selects sweep for the final four FE. The elite set is $\{2,3\}$ in both arms. Select these valid $(i,\ell,r_1,r_2)$ tuples:

\[
(0,2,1,3),\ (1,2,0,3),\ (2,3,0,1),\ (3,2,0,1).
\]

With $F=.5$ and both coordinates crossed, KEEP produces $(.48,.52)$ twice and $(.56,.92)$ twice, with values .0008, .0008, .18, .18. DROP produces $(.50,.50)$ twice and $(.58,.90)$ twice, with values 0, 0, .1664, .1664. The first two candidates are accepted in each arm and the last two rejected. All donor tuples satisfy the distinct-slot restrictions and candidates are inside the box.

At FE 13, the paid historical best is .0008 for KEEP and 0 for DROP, despite the committed point's better local value. The construction uses exact rational arithmetic and an admissible ideal-real random sequence. It establishes a possible finite path, without identifying a NumPy seed or an expectation sign.
\section{A different archive configuration: HXE1}

The HXE1 comparison illustrates why an internal benefit does not fix performance against external policies. H10/S0 main-group ratios were 0.211925 at 10D and 0.118393 at 20D; the prespecified joint gate failed at 10D and passed at 20D. Against L-SHADE, ratios were 0.175405 and 1.851593, with a 10D pass and a 20D failure. Against CMA-ES, H10 reached main targets in 16 of 54 runs across dimensions, compared with 53 of 54 for CMA-ES; the prespecified main-group external-domination stopping condition was triggered.

On the rotated double-well control, H10 reached its target in 8 of 9 runs at each dimension, compared with 5 of 9 for CMA-ES. On the 10D condition-$10^3$ ellipsoid, H10/S0 had a raw-error ratio of 57.7081 although both reached the $10^{-4}$ target in all nine runs. Target-truncated and raw-error analyses describe different precision levels.

\begin{figure}[!htbp]
\centering
\includegraphics[width=.85\linewidth]{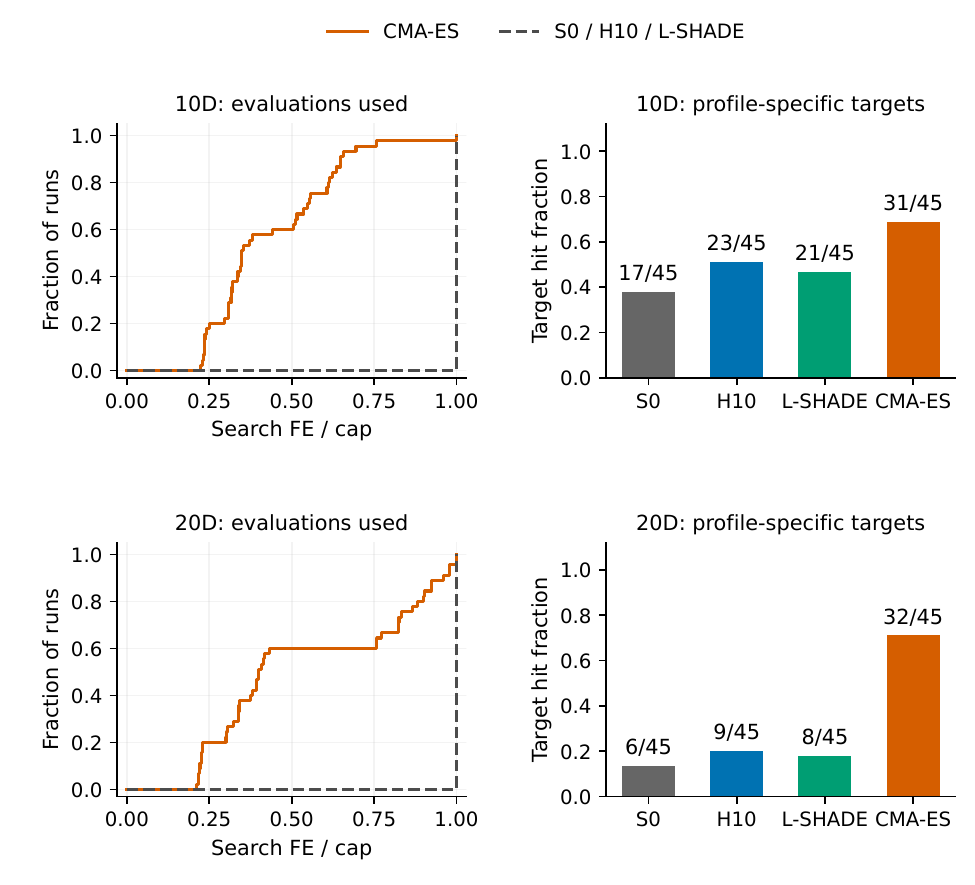}
\caption{HXE1 descriptive search budget use and target hits. Each dimension and method contains 45 runs: five fixed conditions, three instances per condition and three seeds per instance. Left: empirical distributions of actual search FE consumed at termination divided by the FE cap; this is not time to target or wall-clock time. The three non-CMA policies all used the full cap and share the dashed curve; CMA-ES could stop natively without restart. Right: counts of runs meeting the prespecified target for their respective profile. The heterogeneous fixed conditions and repeated seeds do not constitute 45 independent function families. These descriptive counts do not replace the prespecified main-group raw-error comparison.}
\label{fig:external-fe-targets}
\end{figure}

CMA-ES stopped before the cap in 44 of 45 and 43 of 45 runs at 10D and 20D, while the other three methods used the full cap. Figure~\ref{fig:external-fe-targets} shows target hits across the five fixed profiles; these descriptive counts do not replace the original main-group raw-error gate.

\FloatBarrier

\section{Evaluation budgets and stopping}

Every search evaluation counts toward its run's budget, including initialization, rejected candidates, repeated points and incomplete final batches. Verification evaluations occur after search and do not improve its returned solution. Complete-policy comparisons share a maximum budget; a method that stops early retains its best paid value without receiving fictitious evaluations or additional target hits.

In JQX1, each of JQ100, NQ, S0 and L-SHADE used 1,350,000 search evaluations across its 90 runs, whereas CMA-ES used 671,912. In HXE1, each of S0, H10 and L-SHADE used 1,350,000, whereas CMA-ES used 684,808. BX1 and the PC1 native/restart comparison report their actual evaluation totals alongside the results. These totals support comparisons of the specified stopping policies; a common cap alone does not establish equal-cost search efficiency.

\section{Full pseudocode and optional modules}

\begin{figure}[!htbp]
\centering
\includegraphics[width=.92\linewidth]{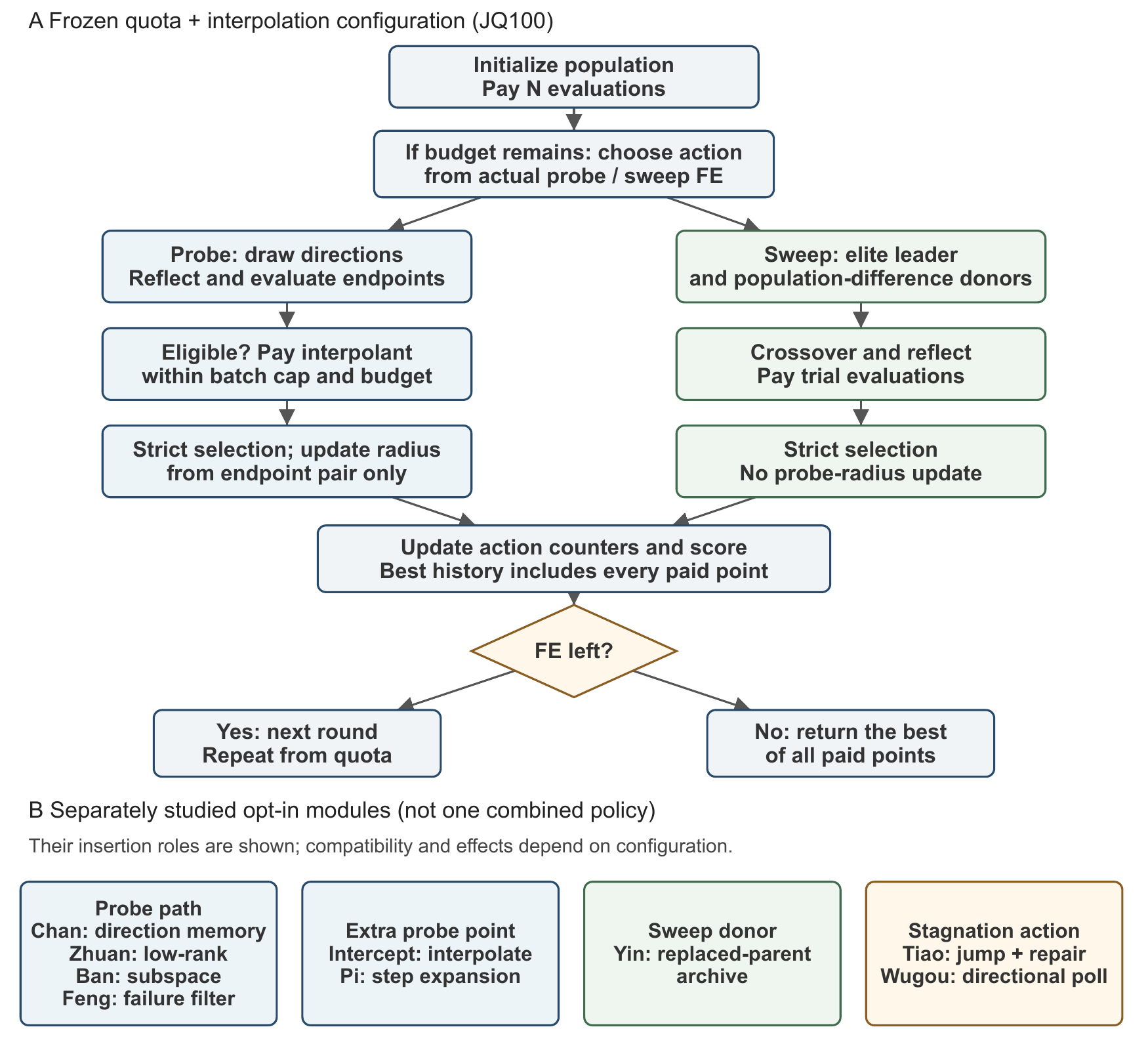}
\caption{DGO workflow and module roles. (A) The fixed JQ100 quota policy studied in J1, J5 and J6: all candidate evaluations, including rejected candidates, enter the budget and best-of-all-paid history; endpoint outcomes alone control probe-radius feedback. NQ disables the interpolation step. (B) Roles of separately studied opt-in modules, shown as a catalog rather than a combined configuration. The integrated research engine restricts compatible combinations: Chan, Zhuan and Ban are alternatives; Feng cannot be combined with Ban or Zhuan; Intercept and Pi are alternatives; and Tiao and Wugou are alternatives. Module behavior and interpolation-fraction semantics depend on the implementation version.}
\label{fig:optional-module-catalog}
\end{figure}

Section~\ref{sec:validity} establishes the finite-budget properties; the listings below specify the fixed policy. Figure~\ref{fig:optional-module-catalog} locates the separately studied optional modules alongside that policy.

\subsection*{Pseudocode for the frozen quota probe--sweep configuration}

The JQ100 configuration used in J1, J5 and J6 assumes a deterministic objective, no scouts, pair-based radius feedback and interpolation cap $\theta\in[0,1]$. KEEP/DROP is a separate experimental intervention.

\begin{lstlisting}[caption={Frozen quota probe--sweep policy with optional interpolation.}]
EVALUATE(P):                         # P contains mapped physical points
    y <- objective(P); nFE <- nFE + length(P)
    update historical best from every (P, y), including rejected points
    return y

INITIALIZE:
    nFE <- 0; historical_best <- infinity
    X <- supplied or uniformly sampled N unit-box vectors
    fitness <- EVALUATE(map_and_clip(X)); r[i] <- 0.1 * sqrt(D)
    cS <- cP <- 0; qS <- qP <- 0.25; clockS <- clockP <- N

WHILE nFE < B:
    pS <- 1 if cS < cP, 0 if cS > cP, else 0.5
    # Short-circuit: no action variate is drawn if only one FE remains.
    probe <- (B - nFE >= 2) AND (Uniform(0,1) >= pS)
    IF probe:
        m <- min(floor(N/2), floor((B - nFE)/2))
        targets <- sample m distinct slots; save their batch-start state
        FOR i in targets:
            u[i] <- normalized Gaussian direction (zero vector -> axis 1)
            delta[i] <- r[i] * u[i]
        endpoints <- reflect(X[targets] +/- delta)
        y_plus, y_minus <- EVALUATE(map_and_clip(endpoints))
        pair[i], y_pair[i] <- better endpoint and its already paid value
                              (break endpoint ties toward plus)
        I_pair[i] <- (y_pair[i] < fitness[i])
        c <- y_plus + y_minus - 2 * fitness[targets]
        alpha <- (y_minus - y_plus) / (2*c) where division is finite
        interpolants <- X[targets] + alpha * delta

        # Eligibility is tested before choosing the target-order prefix.
        eligible <- rows with both unreflected endpoints in the unit box,
                    physical midpoint within tolerance, finite c and alpha,
                    c > 64 * machine_epsilon * (abs(y_plus) +
                         abs(y_minus) + 2 * abs(fitness[i])),
                    1e-8 <= abs(alpha) <= 1 - 1e-8, and mapped candidate
                    distinct from all three mapped nodes
        e <- min(length(eligible), B - nFE, floor(theta * m))
        J <- first e eligible rows in target order  # NQ: e = 0
        IF e > 0:
            yJ <- EVALUATE(map_and_clip(interpolants[J]))
        FOR i in targets, from the saved batch-start population:
            selected[i] <- best of parent and pair[i]
            IF i in J AND yJ[i] < min(fitness[i], y_pair[i]):
                selected[i] <- interpolant[i]
        simultaneously commit strictly improving positions and values
        r[i] <- clip(r[i] * exp(0.7 * (I_pair[i] - 0.5)),
                     1e-12, 0.5 * sqrt(D)) for each target
        cP <- cP + 2*m + e; action <- P; paid <- 2*m + e
    ELSE:
        m <- min(N, B - nFE)
        targets <- all slots if m = N, else sample m distinct slots
        elite <- best min(N, max(2, ceil(0.2*N))) batch-start slots
        leader <- sample an elite for each target, excluding that target
        r1 <- sample donors excluding each target and its leader
        r2 <- sample donors excluding each target, leader and r1
        F <- draw m independent Uniform(0.4, 0.9) factors
        v <- X[targets] + F * (X[leader] - X[targets] + X[r1] - X[r2])
        mask <- draw independent Bernoulli(0.9) for each coordinate
        force one crossover coordinate per target
        trial <- where(mask, v, X[targets])
        y <- EVALUATE(map_and_clip(reflect(trial)))
        simultaneously commit strictly improving trials and values
        cS <- cS + m; action <- S; paid <- m
    q[action] <- 0.8*q[action] + 0.2*number_committed/paid
    clock[action] <- nFE            # scores do not choose quota actions
RETURN best position and value among all paid points
\end{lstlisting}

\noindent\begin{minipage}{\linewidth}
\subsection*{Pseudocode for a single commitment intervention}

\begin{lstlisting}[caption={One-commitment intervention used in the J5 and J6 DGO studies.}]
Follow the prescribed carrier; select its specified improving third point
Save the settled round-end state and candidate evaluation record, including
    population, values, radii, FE counters, scores, clocks, history,
    best-of-all-paid and both random-generator states
KEEP <- unmodified saved state
DROP <- deep copy of saved state
IF the selected endpoint pair improved its parent:
    DROP.selected_slot <- the successful best endpoint
ELSE:
    DROP.selected_slot <- the saved parent
    decrement probe-success count and recompute the probe score from the
        pre-round score and the actual post-intervention successes
Retain the third-point cost and observation, historical best, radius,
    logical FE, allocation counters, other slots and both RNG states
Retain the recorded potential third-point-improvement count; record the
    suppressed population commitment separately
Check coordinate--fitness consistency and identity outside allowed changes
Replay KEEP with the original streams and verify the carrier suffix exactly
IF the recovery gate passes:
    J5: continue KEEP and DROP with the original streams to B
    J6: continue KEEP and DROP with each prespecified paired new stream to B
    Do not force later actions, donors or random-number consumption to match
\end{lstlisting}
\end{minipage}

\FloatBarrier
\section{Detailed experimental contracts}

\subsection{Custom test functions}

The J1–J3 and J6 functions use a physical box $[-5,5]^D$, shifts $o\in[-3,3]^D$, and a QR-derived orthogonal matrix $Q$ for rotated conditions. Set $z=x-o$ for axis-aligned functions and $z=(x-o)Q$ for rotated ones, using the implementation's row-vector convention. For $D>1$, the sphere, ellipsoid and Rastrigin families are

\[
f_{\rm sphere}=\sum_i z_i^2,\quad
f_{\rm ellip}=\sum_i \kappa^{(i-1)/(D-1)}z_i^2,\quad
f_{\rm rast}=\sum_i[z_i^2+10(1-\cos(2\pi z_i))].
\]

For Rosenbrock, set $u_i=z_i+1$ and use $\sum_{i=1}^{D-1}[100(u_{i+1}-u_i^2)^2+(u_i-1)^2]$. The Ackley implementation uses $\max(0,-20e^{-0.2\sqrt{\operatorname{mean}(z_i^2)}}-e^{\operatorname{mean}(\cos(2\pi z_i))}+20+e)$. The J1–J3 quadratic group contains sphere, axis and rotated ellipsoids with $\kappa=10^6$, and a rotated ellipsoid with $\kappa=10^2$; axis and rotated Rosenbrock, Ackley, and rotated Rastrigin form the other fixed groups. Shared geometry within an instance block is a pairing feature, not a new independent function family.

HXE1 uses a separate, previously fixed suite with shifts in $[-0.8,0.8]^D$. Its rotated main group comprises ellipsoids with $\kappa=10^3$ and $10^6$ and Rosenbrock. Its controls are rotated Rastrigin and $\min(\|z\|^2,0.2+\|z-2e_1\|^2)$. The respective main/control targets are $10^{-4},10^{-2},10^{-3},10^{-2},10^{-6}$. Fixed problem arrays, seeds, implementation versions and exact experiment contracts remain part of the reproducibility record; these custom functions are not official benchmark instances.

A base run supplies the states used for diagnostic branches or interventions. J2 captures states after the endpoint evaluations and before the third-point and radius updates. J is the interpolation point and N is a new-direction point; JN continues from the accepted center after J, whereas NN evaluates a new-direction pair. Random-interior and pre-reflection equal-length controls were auxiliary. In J3, I uses the old interpolation point first, E+ evaluates a fresh pair and may reuse its new interpolation model, and E0 disables reuse of that first model. J5 deliberately selected one tail seed and one seed that benefited relative to NQ. Its DROP state restores the successful endpoint or the parent while retaining the paid point and history as specified in the intervention pseudocode.

HXE1's H10 archive of replaced parents was initially empty. S0/H10 used 40 individuals; L-SHADE retained native population reduction from 18D to four and parent-midpoint repair, while CMA-ES used the fixed pycma 4.5.0 adapter and native stopping without restart.

S0 and H10 shared an initial population and search seed but not necessarily their complete later random streams. The fixed L-SHADE adapter used the author's corrected 1.0.1 source, a success-history size of five, archive rate 1.4 and $p=0.11$; it repaired an out-of-box coordinate to the midpoint between its parent and the violated boundary. Its strict-budget bridge evaluated only the remaining prefix of a final generation and disabled the original small-error zeroing. CMA-ES used an independently sampled unit-box mean in $[0.2,0.8]^D$, initial scale 0.3 and the native BoundTransform. A final incomplete CMA generation was evaluated only up to the remaining budget and not passed to tell, but all paid points contributed to the shared historical-best output. The common cap does not equalize initialization, actual FE or historical tuning budgets.

The HXE1 main-group ratio is the exponential of the difference in mean $\log(\max(h,10^{-12}))$ between the two policies, averaging equally over seeds, then instances, then the three main conditions. The design is balanced at these levels. It is a geometric-error ratio for that fixed mixture, not a ratio of arbitrary raw-mean risks or an estimate for an unspecified population of problems. Target-hit counts are a separate descriptive outcome with the condition-specific targets defined above.

\subsection*{Decision thresholds}
The numerical thresholds encode design decisions: a ratio of 0.9 requires a 10\% reduction on that ratio scale, 1.25 permits a 25\% profile increase, and a worst-case ratio of 10 limits a severe matched-case loss. Their joint use defines a study-specific screen; the thresholds were not estimated as optimal or validated as universal practical-loss boundaries.

\subsection{Prespecified classical and random-reference comparison: JQC1}

The JQC1 study compares the original JQ100 policy, also used in J1, J5, J6 and JQX1, with three fixed references. JQ100 uses 40 individuals, a 0.5 evaluation quota for sweep, no scout, pair-based radius feedback with $\eta=0.7$, initial radius $0.1\sqrt D$, and full eligible interpolation fraction; its original sweep settings are retained. DE/rand/1/bin uses $N=5D$, $F=0.5$, $CR=0.9$, three distinct donors excluding the target, forced crossover and synchronous strict selection \citep{storn1997de}. Positions are reflected into the box. The synchronous global-best PSO uses 40 particles, zero initial velocity, independent coordinatewise random factors, $\phi_1=\phi_2=2.05$, $\phi=4.1$ and
\[
 v^{+}=\chi\{v+\phi_1 r_1(p-x)+\phi_2 r_2(g-x)\},\qquad
 \chi=\frac{2}{\phi-2+\sqrt{\phi^2-4\phi}}.
\]
Personal bests require strict improvement and global-best ties use the lowest slot. Positions use triangle-wave reflection and velocity signs follow reflection parity, pointing inward on an exact boundary; there is no velocity cap or restart. This specified implementation uses Clerc--Kennedy coefficients \citep{clerc2002particle}; it is not SPSO-2011, and no convergence theorem is transferred to the reflected policy. Random search uses independent uniform points after the initial 40. All methods return the best of all paid observations, including initialization, rejected points and tails.

JQ100's interpolation fraction was selected in J1 from 0.25, 0.50 and 1.00; DE and PSO had no corresponding JQC1 tuning stage. The comparison concerns fixed policies on fresh instances, with unequal historical selection budgets.

The fixed design specifies eight custom profiles in six families at $D=10,20$, three fresh instances per profile and dimension, and five nested search seeds, for 960 runs and budgets 10,000 and 20,000 evaluations per run. A master $5D$-point initial array supplies the same first 40 points to JQ100, PSO and Random; DE receives all $5D$ points and pays for its additional initial evaluations. Thus DE does not have the same initial-best distribution, and matching seed labels do not imply common random numbers. The suite is not an official CEC or BBOB benchmark, and independent axis and rotation instances do not isolate rotation causally.

The profiles are Sphere, axis-aligned and rotated ellipsoids with condition number $10^6$, rotated Rosenbrock, axis-aligned and rotated Rastrigin, rotated Ackley and rotated doublewell. Targets are respectively $10^{-8}$, $10^{-2}$ for both ellipsoids, $10^{-3}$, $10^{-2}$ for both Rastrigin profiles, $10^{-3}$ and $10^{-6}$. Each instance uses the box $[-5,5]^D$, a shift sampled uniformly from $[-0.8,0.8]^D$ and a QR-derived rotation where specified. The doublewell objective is $\min(\|y\|^2,0.2+\|y-2e_1\|^2)$ in shifted, rotated coordinates. All optima are zero; target values are fixed before any run.

The primary error is $\max(f-f_*,\tau)$ for each profile target $\tau$; $\max(f-f_*,10^{-12})$ is auxiliary. Within each profile, paired log ratios are averaged across seeds and instances; the two ellipsoid profiles and the two Rastrigin profiles are averaged within their families, then six families have equal weight. Dimensions are reported separately.

Against Random, the prespecified basic screen requires, in each dimension, a family-macro ratio at most 0.9, at least five of six family ratios at most 1, no family ratio above 1.25, and no decrease in family-weighted target-hit rate. Both dimensions are required for the overall basic screen. Separately against DE and PSO in each dimension, the classical parity screen requires macro ratio at most 1.25, at least four of six family ratios at most 1.25, none above 10, and no lower family-weighted hit rate; the stronger improvement label additionally requires macro ratio at most 0.9 and at least four family ratios below 1. These are descriptive study-specific rules, not general definitions of optimizer usability.

Prespecified 2,000-replicate hierarchical paired bootstrap intervals resample three instances within each profile, then five paired seeds within sampled instances; they are conditional on these fixed families and are not multiplicity-adjusted population-of-functions inference. All 960 runs completed, using 14,400,000 search evaluations.

\subsection{New-state intervention: J6}

J6 used four custom conditions, dimensions 10 and 20, six new instances per condition and dimension, one base-run seed per instance and a budget of $1000D$ FE. The rotated ellipsoids with condition numbers $10^2$ and $10^6$ were the primary quadratic group; rotated Rosenbrock and Rastrigin were prespecified controls. Each base run could yield at most one checkpoint: the first accepted eligible third point in the first qualifying batch within the prespecified 20--40\% budget window. The exposure denominator included base runs without such a point.

After an exact original-stream KEEP recovery gate, each selected state was continued as KEEP and DROP under four paired new random streams. The study recorded parent and local-best improvements and evaluated 16 one-point new-direction controls and 16 random interior controls without feeding those branches back to either replay arm. For each independent normalized standard-Gaussian direction $u$, the reference point is $x_N=\operatorname{map}(\operatorname{reflect}(x_0+\|\delta\|u))$, where $x_0$ is the selected probe's original parent in unit-box coordinates and $\|\delta\|$ is its original displacement length before radius feedback. With $f_b$ as the paid parent-and-endpoint minimum, $m_N=\min(f_b,f(x_N))$ and $m_J=\min(f_b,f_j)$. This reference differs from J2's continuation from the paid-best point using its post-feedback radius. The normalized local estimate was
\[
\widehat L_{\rm new}=\frac{\overline m_N-m_J}{\max(|f_b|,10^{-12})}.
\]
The local 16 replicates estimate an average one-point comparison, not a 16-point search policy. Each saved state record includes the four-pair mean raw DROP-minus-KEEP difference. The tables and joint criterion below use the distinct estimator of $W_B$,
\[
\widehat v_{\log}=\frac14\sum_{r=1}^{4}
\ln\left(\frac{h_{D,r}+10^{-12}}{h_{K,r}+10^{-12}}\right).
\]
A state estimate met the joint criterion when $\widehat L_{\rm new}\ge0.01$ and $\widehat v_{\log}\le\ln(0.9)$; the opposite material-effect threshold was $\widehat v_{\log}\ge\ln(1.1)$. These empirical statistics use the nonnegative, zero-optimum objectives specified above and $\epsilon=10^{-12}$. Adding a constant to an objective can change the normalization, log ratio and threshold classification, even if the candidate path is unchanged; near the numerical floor their interpretation also depends on $\epsilon$. They are not translation-invariant criteria for arbitrary real-valued objectives. The four future pairs are repeats within a state, not four problem instances. Thresholding their mean does not establish the sign of a state's exact expected effect on either scale.

We report 95\% percentile intervals from the prespecified 2,000 bootstrap resamples, which resampled instances within condition and dimension and paired continuations within resampled states; local references were also resampled where required for joint quantities. Functions and dimensions remained fixed strata. The fixed research-stage transfer gate required at least two distinct instances meeting the joint criterion and having KEEP worse in at least three of four paired continuations. This gate selected a transfer question and did not establish a general effect or an online policy.

\subsection{Controlled second-skeleton intervention: M3}

M3 used 32 new base runs: four new instances in each of the same four conditions at each dimension, with one seed per instance. The base run used an independently implemented synchronous DE/rand/1/bin policy with $N=40$, $F=0.5$, crossover probability 0.9, distinct donors excluding the target, reflected box boundaries and strict selection; it used no archive, quota, elite guidance or adaptive slot radii.

A single diagnostic probe batch was inserted after the completed generation at 30\% of $B=1000D$. The batch selected 20 targets, paid two reflected endpoints per target at a fixed unit-box radius $0.1\sqrt D$, and paid all eligible interpolation points under the fixed interpolation guards. It then committed the best available candidate per target. The first strictly better eligible third point in the prescribed target order was selected at batch end; the choice did not use the local reference outcomes or futures. A base run without such a candidate retained its denominator and received no replacement probe. This policy is DE/rand/1/bin plus one diagnostic probe, not unmodified DE.

The KEEP/DROP treatment retained the paid third point, history, logical FE and all other slots; DROP restored the selected slot to its successful endpoint or parent, with required non-decision bookkeeping. The original and captured base runs and original-stream KEEP suffix first had to agree exactly. Then four paired fresh DE streams continued each selected state to $B$. The 16 fresh-direction single-point references used the original selected parent and the same fixed probe radius with the original reflection rule, and were not fed back to either arm. Their normalized local contrast and the four-future log contrast use the same definitions and thresholds as J6; there was no random-interior commitment comparison.

Within each condition and dimension, four instances were resampled with replacement, then four paired futures and 16 local reference values were resampled within selected states. The prespecified 2,000 stratified bootstrap resamples recomputed the estimates and classifications; eligible-state mean log ratios gave equal weight to the conditions. Each primary condition-dimension cell required at least three of four selected states for sufficient exposure. Base runs without a selected state remained in the denominators for counts.

If a resample contains no selected state in a condition, the equal-condition mean is unestimable; the reported interval is conditional on the resamples in which this mean can be computed. The stricter case criterion combined the two local and log thresholds with KEEP being worse in at least three of four paired futures. Neither this criterion nor the confidence intervals establish individual state expectation signs.

\subsection{Same-version external comparison: JQX1}

JQX1 was fixed after the mechanism studies to compare the same JQ100 policy used in J1, J5 and J6 with complete solvers. It uses the unchanged JQ100 implementation, including the initial radius $0.1\sqrt D$, pair-based radius feedback, quota allocation, full third-point fraction and no scout action. NQ uses the identical core with interpolation disabled. S0 is the fixed pure-sweep policy with no archive. The L-SHADE and CMA-ES versions, native initialization, boundary handling and stopping contracts are the same as in HXE1; no method receives new tuning or restarts. JQ100, NQ and S0 share a 40-point initial population for each case, whereas external solvers retain their own initialization. Shared seed labels therefore do not mean identical starting populations or semantic random choices across all five methods.

At each of 10D and 20D, the study uses three fresh instances and three search seeds for each of five rotated conditions, with caps of 10,000 and 20,000 FE, respectively. This gives 450 runs across five methods. The two prespecified primary conditions are ellipsoids with $\kappa=10^2$ and $10^6$; rotated Rosenbrock, Rastrigin and double-well functions are stress controls. Functions and the $[-5,5]^D$ box follow the HXE1 suite, with fresh shifts, rotations and starts checked against archived inputs. Their respective targets are $10^{-4}$, $10^{-2}$, $10^{-3}$, $10^{-2}$ and $10^{-6}$. The double-well starts are uniform, not deliberately trapped. These are new instances of fixed custom families, not an official benchmark suite or unseen-family validation.

Within each dimension, geometric final-error ratios use a $10^{-12}$ floor and equal seed, instance and condition weights. Against each of NQ, S0, L-SHADE and CMA-ES, the primary gate requires a JQ100/reference ratio at most 0.9, no fewer target hits, every condition ratio at most 1.25, the worst matched case ratio at most 10, and a ratio below one in at least two of the three predefined instance blocks. Stress controls require the latter condition-wise and worst-case limits and no fewer total hits. Overall support requires both dimensions to pass all internal, external and control gates; primary-only support has a separate specialist label. These are prespecified descriptive practical-effect gates, not significance or equivalence tests.

The strong external-domination criterion additionally requires consistent condition and instance-block losses with no primary hit or paired target-speed advantage. Failure to establish that strong criterion does not count as passing the comparison. Every scientific outcome is retained; completion stops the study without adding seeds, tuning or modules.

\subsection{Retrospective descriptive analysis}

The P0 analysis reprocessed saved J1, J2, J3, J5 and HXE1 outputs without new optimization or objective evaluations. J1 used $\log(f+10^{-12})$; HXE1 used $\log(\max(f,10^{-12}))$. Ratios are candidate/reference, with smaller values favoring the candidate. For descriptive sensitivity, 20,000 hierarchical paired bootstrap resamples used seed 2026102901; J1 resampled shared geometric instance blocks jointly across conditions, whereas HXE1 resampled actual instances within condition. Leave-one-instance analyses and fixed target-truncation summaries were retained. With only three instances per condition, repeated prior inspection and no multiplicity adjustment, the intervals do not retroactively create confirmatory significance.

\section{Selected historical commitment cases}

The J1 specialist comparison and the local-branch studies use different samples and are not pooled; their juxtaposition motivates a separate intervention.

J1's held-out quadratic group favored JQ100 over NQ and RQ100 under the original hierarchical geometric aggregation: ratios were 0.179529 and 0.0590668, respectively, and both prespecified specialist gates passed. JQ100 did not meet the broad default gate. On one rotated condition-$10^6$ ellipsoid instance and search seed, JQ100/NQ was 223.363; the corresponding three-seed instance-block geometric ratio was 6.11726. Aggregate specialist support coexisted with a severe observed tail case.

The P0 descriptive 95\% percentile interval for the quadratic JQ100/NQ aggregate was [0.129586, 0.269004], with three shared instance blocks and search-seed repetitions. This retrospective interval must be read with the individual tail rather than as a new confirmatory test. In a fixed target-truncation sensitivity analysis, the same aggregate ratio changed from 0.179529 on the original scale to 0.858310 at $10^{-4}$ and 0.369163 at $10^{-8}$, reflecting different weights on refinement after reaching those targets.

The preselected J5 tail case had terminal values 3.310171365 for KEEP and 0.522353440 for DROP, a DROP/KEEP ratio of 0.157803. Suppressing this one commitment improved the recorded continuation, but DROP still ended at 35.247 times the old NQ value. In the reference case, KEEP ended at 0.008463288 and DROP at 1.391169817, giving a ratio of 164.377. The same intervention type produced opposite effects on these specified trajectories. Neither case isolates a donor-, radius- or schedule-specific direct effect.

Retrospective reconstruction located the first candidate-coordinate differences in round 3 in both cases. The first elite/donor-identifier, radius and action differences appeared in rounds 7/8/16 for the tail and 3/4/27 for the reference. Temporal order describes divergence; after divergence, equal round numbers do not denote equal states. Donor-coordinate differences can occur even when identifiers are unchanged.

\section{Supplementary profile and state displays}

\begin{figure}[!htbp]
\centering
\includegraphics[width=\linewidth]{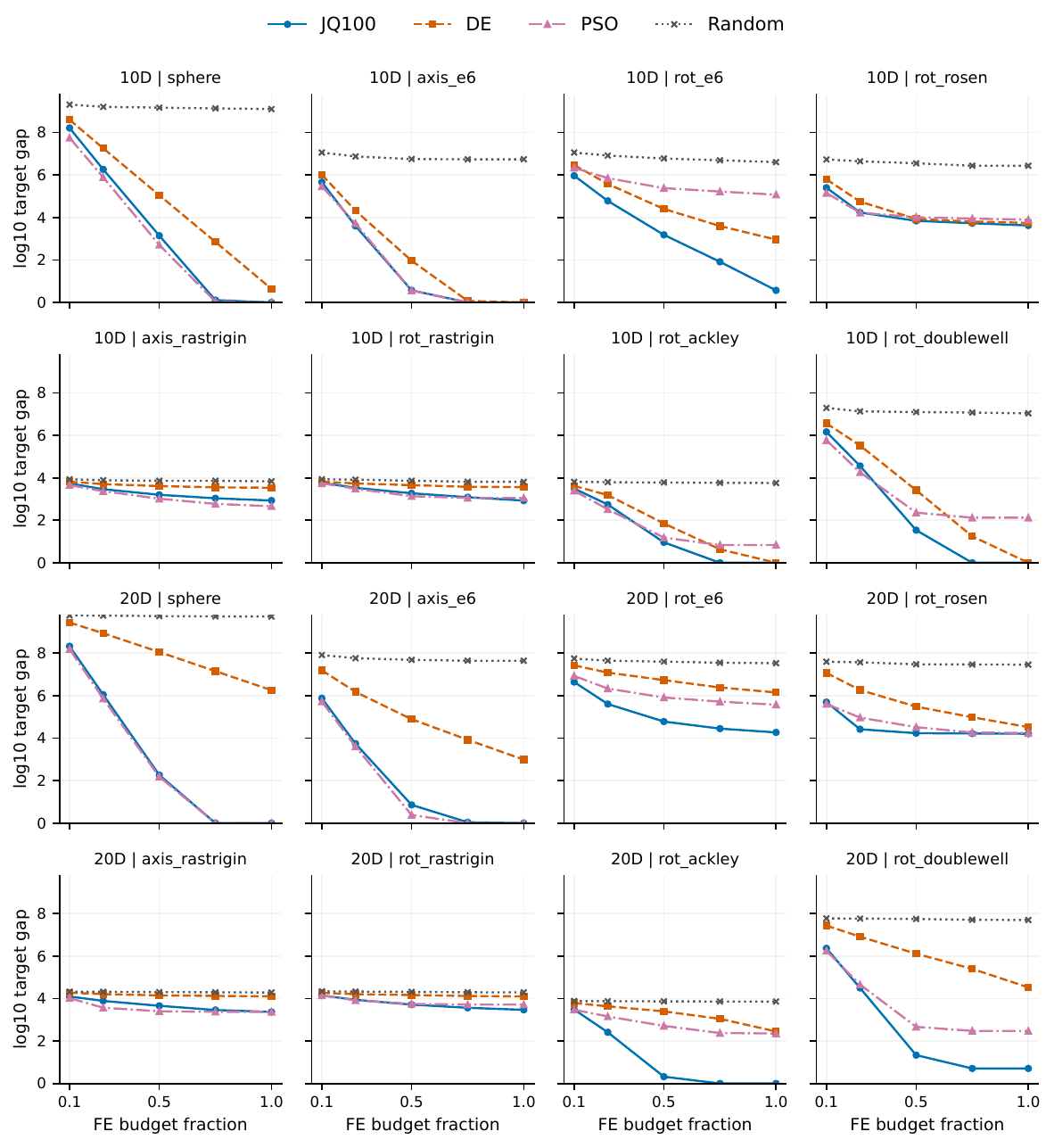}
\caption{JQC1 paid-batch trajectories by profile and dimension. Each curve displays the saved geometric target-truncated error divided by the profile target on a $\log_{10}$ scale, over three instances and five seeds. Curves use completed batches at the indicated FE clocks. These are descriptive profile averages, with no new confidence band or statistical resampling.}
\label{fig:classic-profiles}
\end{figure}

\begin{figure}[!htbp]
\centering
\includegraphics[width=\linewidth]{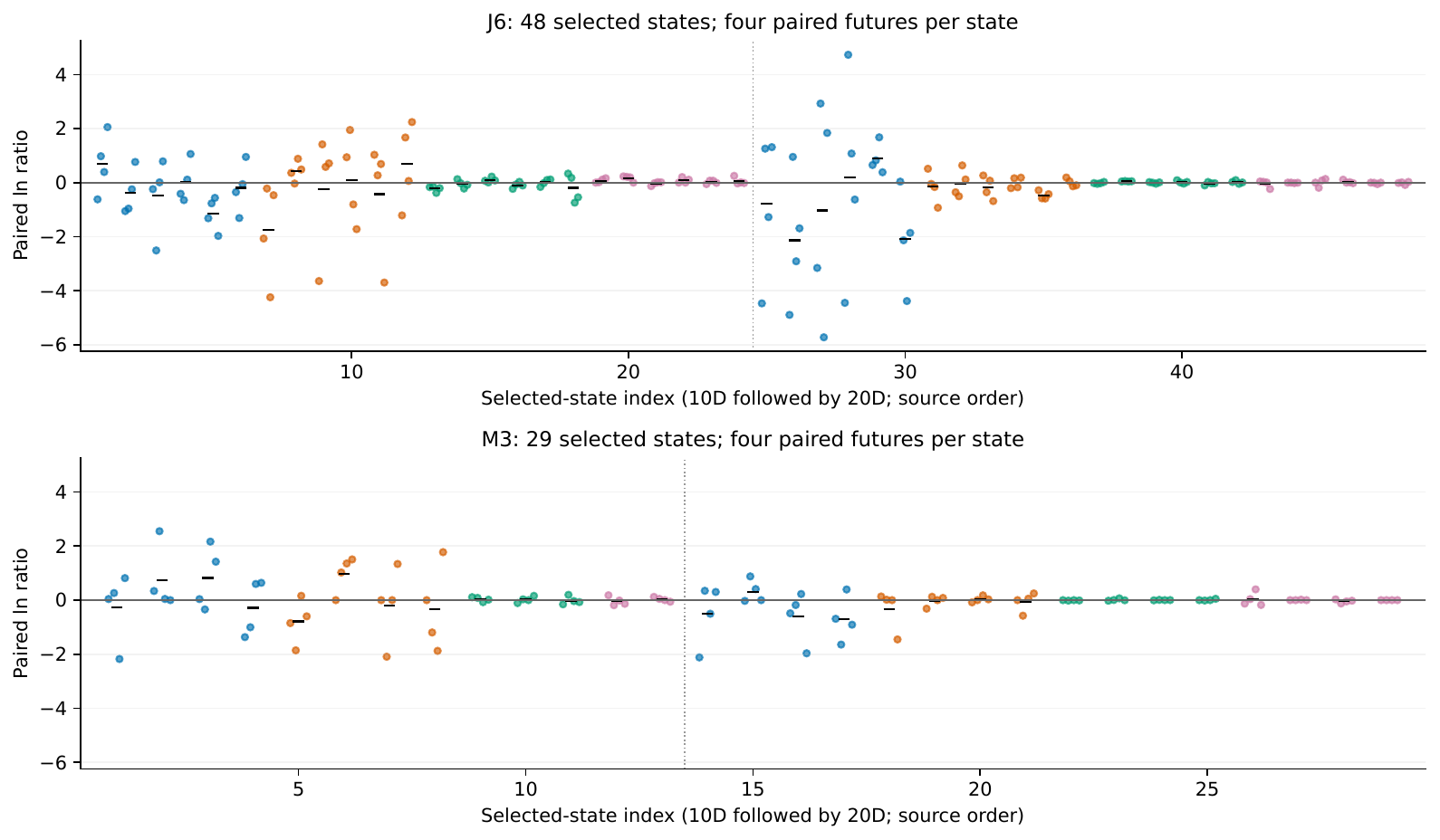}
\caption{Saved paired continuation outcomes for the 48 selected J6 and 29 selected M3 states. Colored dots show the four values of $\ln[(h_D+10^{-12})/(h_K+10^{-12})]$ within each state; black ticks show their saved mean. Small horizontal offsets separate the four repeats without changing state membership. The vertical dotted line separates 10D and 20D states, and the horizontal line marks zero. States follow the source-record order; four futures are within-state repeats.}
\label{fig:paired-states}
\end{figure}

\begin{figure}[!htbp]
\centering
\includegraphics[width=\linewidth]{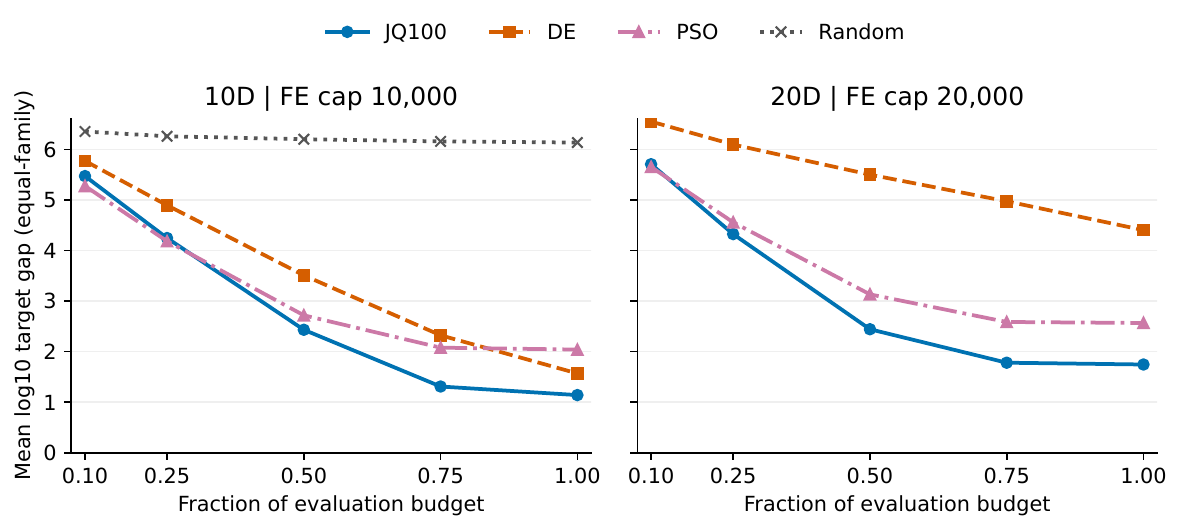}
\caption{Finite-budget trajectories for the four fixed JQC1 policies. Each panel averages $\log_{10}(\max(e,\tau)/\tau)$ within profiles and then weights the six families equally. Values at a budget clock use completed paid batches only; no partial batch is treated as observed.}
\label{fig:classic-baselines}
\end{figure}

The profile trajectories in Figure~\ref{fig:classic-profiles} show the individual conditions underlying the family-macro trajectory in Figure~\ref{fig:classic-baselines}. Figure~\ref{fig:paired-states} displays each saved paired continuation and its state mean for the selected J6 and M3 states.

\FloatBarrier
\subsection*{Full comparison tables and supporting displays}
Table~\ref{tab:study-map} summarizes the earlier custom-function experiments, and Tables~\ref{tab:j6}--\ref{tab:m3} provide the corresponding cell-level continuation summaries. These supplement the principal comparisons in Results; CT1, PC1 and the official boxed benchmark are specified in Appendix~\ref{app:followup}.

\begin{table}[!htbp]
\centering
\small
\caption{Study designs and the units to which their comparisons apply. These earlier studies use custom functions; HXE1 and JQX1's external solvers are L-SHADE and CMA-ES \citep{tanabe2014lshade,hansen2023cma}.}
\label{tab:study-map}
\begin{tabular}{>{\raggedright\arraybackslash}p{1.8cm}>{\raggedright\arraybackslash}p{6.7cm}>{\raggedright\arraybackslash}p{5.5cm}}
\toprule
Study & Comparison and sampled units & Scope of interpretation \\
\midrule
JQC1 & At 10D and 20D: eight profiles in six families, three fresh instances and five seeds per profile, four policies; 960 runs. & Unchanged JQ100 versus fixed DE, PSO and Random; common paid budget, shared 40-point initialization prefix, six-family weighting. \\
J1-QS & 10D; eight conditions; three new instance blocks and three seeds per condition. Five policies; 360 held-out runs. & Complete policies with their own later trajectories. \\
J2-OC & 96 fresh base runs, 5,760 paid-endpoint states, 5,194 eligible; eight repetitions. J versus N costs one FE; JN versus NN costs two FE. & Local branches do not update the base run; average within state before equal-weight hierarchical log aggregation. \\
J3-DV & 1,440 fixed-rule selected fresh states; four paired repetitions of three 24-FE policies. The 2-, 4- and 8-FE prefixes were prespecified. & Local policies do not change the full population or share discoveries. \\
J5-PATH & Two selected J1 seeds on one rotated condition-$10^6$ ellipsoid instance; first accepted point at round 2, FE 121; continuation to 10,000 FE. & Original-stream KEEP recovery required; illustrative cases, not new held-out samples. \\
J6 & 48 selected states from 48 base runs; four paired continuations per state, 16 fresh-direction and 16 random-interior one-point references. & Same-candidate local and commitment estimates; futures are nested within states. \\
M3 & 29 selected states from 32 base runs across dimensions; four paired continuations per selected state and 16 one-point local references. & Controlled DE plus probe; retain unselected base runs in exposure denominators. \\
HXE1 & At 10D and 20D: three main conditions, two controls, three new instances, three seeds and four algorithms; 360 runs. & S0 and archive-based H10 versus external solvers under a common cap, with different initialization and native stopping. H10 has no JQ100 interpolation. \\
JQX1 & At 10D and 20D: two quadratic primary conditions, three stress controls, three fresh instances, three seeds and five policies; 450 runs. & The same fixed JQ100 as the mechanism studies, versus NQ, S0, L-SHADE and CMA-ES; fixed configuration, common cap and native external stopping. \\
\bottomrule
\end{tabular}
\end{table}

\begin{table}[!htbp]
\centering
\small
\caption{J6 selected-state and continuation summaries. Mean log ratios and percentile intervals use natural logarithms. Joint counts state estimates with $\widehat L_{\rm new}\ge0.01$ and $\widehat v_{\log}\le\ln(0.9)$; the count is not the number of confirmed true state effects. Hits are KEEP and DROP target hits across four paired continuations per selected state, so their denominators are nested within states. All combines the two quadratic and two control conditions; it is not a sample of independent function families.}
\label{tab:j6}
\begin{tabular}{lcccc}
\toprule
Group & Selected & Mean log ratio [95\% interval] & Joint & Hits K, D \\
\midrule
10D quadratic & 12/12 & $-0.223\;[-0.764,\,0.279]$ & 3/12 & 25/48, 24/48 \\
10D all & 24/24 & $-0.116\;[-0.399,\,0.130]$ & 5/24 & 25/96, 24/96 \\
20D quadratic & 12/12 & $-0.484\;[-1.131,\,0.153]$ & 3/12 & 9/48, 10/48 \\
20D all & 24/24 & $-0.242\;[-0.575,\,0.081]$ & 3/24 & 9/96, 10/96 \\
\bottomrule
\end{tabular}
\end{table}

\begin{table}[!htbp]
\centering
\small
\caption{M3 selected-state and continuation summaries. Means and percentile intervals use natural logarithms, average eligible states within condition and weight conditions equally. Joint counts use the J6 two-threshold definition; stricter counts additionally require KEEP worse in at least three of four paired futures. Count denominators retain all base runs, including those without selected states. All combines the two primary and two control conditions. For the 10D all-condition mean, 145/2,000 bootstrap draws were unestimable; its interval uses the remaining 1,855 draws. Four futures per state do not confirm individual expected signs.}
\label{tab:m3}
\begin{tabular}{lcccc}
\toprule
Group & Selected & Mean log ratio [95\% interval] & Joint & Stricter \\
\midrule
10D quadratic & 8/8 & $0.085\;[-0.464,\,0.647]$ & 4/8 & 1/8 \\
10D all & 13/16 & $0.044\;[-0.225,\,0.304]$ & 4/16 & 1/16 \\
20D quadratic & 8/8 & $-0.235\;[-0.537,\,0.049]$ & 4/8 & 2/8 \\
20D all & 16/16 & $-0.117\;[-0.271,\,0.021]$ & 4/16 & 2/16 \\
\bottomrule
\end{tabular}
\end{table}

\begin{table}[!htbp]
\centering
\small
\caption{JQX1 quadratic primary comparisons. Each dimension/method has 18 runs: two fixed conditions, three instances and three seeds. Ratios use the $10^{-12}$ error floor and equal hierarchical weights; smaller favors JQ100. Hits are JQ100/reference, each out of 18. The gate also checks each condition, predefined instance blocks and the worst matched case.}
\label{tab:jqx1}
\begin{tabular}{llrrrl}
\toprule
D & Reference & Error ratio & Hits & Worst ratio & Criteria \\
\midrule
10 & NQ & 0.204 & 10/9 & 5.259 & Met \\
10 & S0 & 104.5 & 10/11 & $1.009\times10^{5}$ & Not met \\
10 & L-SHADE & 3.673 & 10/11 & 64.68 & Not met \\
10 & CMA-ES & $4.142\times10^{7}$ & 10/18 & $5.726\times10^{11}$ & Not met \\
20 & NQ & 1.536 & 6/7 & 58.41 & Not met \\
20 & S0 & 0.3757 & 6/3 & 13.84 & Not met \\
20 & L-SHADE & 8.185 & 6/9 & 137.6 & Not met \\
20 & CMA-ES & $9.080\times10^{7}$ & 6/15 & $3.554\times10^{14}$ & Not met \\
\bottomrule
\end{tabular}
\end{table}

\begin{figure}[!htbp]
\centering
\includegraphics[width=\linewidth]{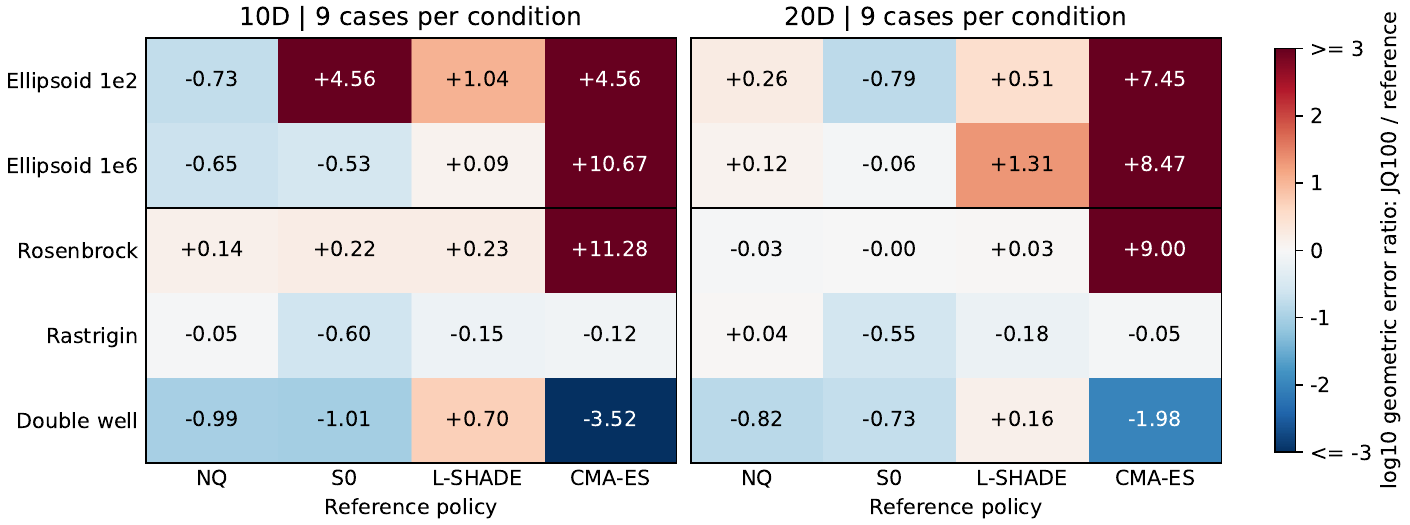}
\caption{JQX1 complete-policy comparisons on every prespecified condition. Entries are $\log_{10}$ geometric final-error ratios of JQ100 to each reference, with a $10^{-12}$ numerical floor; negative values favor JQ100. Each cell summarizes three instances and three seeds. Colors saturate at $\pm3$, while annotations retain the actual log ratios. The line separates the two quadratic primary conditions from the three stress controls.}
\label{fig:jqx1}
\end{figure}

\FloatBarrier
\section{Follow-up designs and supporting results}\label{app:followup}

The follow-up comprised three separately prespecified experiments and one saved-data analysis. CT1 addressed candidate type, state-population precision and radius adaptation; PC1 addressed initialization, restart, parameter sensitivity and implementation cost; BX1 tested public benchmark transfer. S1 reanalysed 2,026 saved trajectories from JQC1, JQX1, J6 and M3. No configuration was selected from the follow-up outcomes.

\subsection{Preselected-state commitment: CT1}
The four profiles were rotated ellipsoids with condition numbers $10^2$ and $10^6$, rotated Rosenbrock and rotated Rastrigin. Each profile had 16 new instances at each of 10D and 20D, giving 128 independently generated instances and one NQ state per instance. With $N=40$ and $B=500D$, each base run stopped after the complete round at $T=40\lfloor0.3B/40\rfloor$, namely 1,480 or 3,000 FE. Slot and candidate streams were specified before observing eligibility or candidate outcomes.

At the selected parent and current radius, the diagnostic paid for two reflected endpoints and four candidate evaluations. I used the prespecified interpolation guards and reverted to the parent if ineligible; R sampled a signed interior fraction on the original direction; N used a fresh normalized Gaussian direction at the same radius; E reevaluated the better original endpoint. Repeated endpoints and the ineligible-I parent were charged. All six values entered a common paid history. Only a type's strict parent improvement was committed in its KEEP arm; DROP kept the parent. Radii and scores were unchanged by the diagnostic itself. Thus nonaccepting types produced identical KEEP/DROP states and remained in the all-state analysis. Continuations used NQ to budget $B$, with two paired streams per state.

For interpolation, a second KEEP/DROP pair used the same futures with all 40 slot radii fixed at their valid checkpoint values. The contrast compares total commitment effects under natural and fixed-radius continuation policies; it is not a natural mediation decomposition. There were 2,560 continuations, plus 128 base-run prefixes and 128 diagnostic batches. Table~\ref{tab:ct1-details} reports all quadratic-group CT1 effects and paired contrasts.

\begin{table}[!htbp]
\centering\small
\caption{CT1 quadratic group: 64 preselected states and two futures per state. Entries are means and 95\% pointwise percentile intervals for log DROP/KEEP or paired differences of that quantity. Positive values favor KEEP for a single type; a positive contrast indicates a larger commitment effect for its first type or radius policy.}
\label{tab:ct1-details}
\begin{tabular}{lrr}
\toprule
Quantity & Mean & 95\% interval \\
\midrule
Interpolation I & -0.04023 & [-0.16865, 0.09088] \\
Random interior R & -0.01512 & [-0.08134, 0.04642] \\
New direction N & -0.00667 & [-0.06962, 0.05181] \\
Better endpoint E & 0.01610 & [-0.03702, 0.06556] \\
I with fixed radii & 0.03336 & [-0.11766, 0.19132] \\
I minus R & -0.02511 & [-0.15232, 0.10693] \\
I minus N & -0.03356 & [-0.16155, 0.09219] \\
I minus E & -0.05633 & [-0.18663, 0.07257] \\
I natural minus fixed & -0.07359 & [-0.28428, 0.13335] \\
\bottomrule
\end{tabular}
\end{table}

Interpolation was eligible in 110 of the 128 states. Acceptance counts and the unchanged accepted set under the stricter endpoint rule are reported in Table~\ref{tab:ct1-acceptance}. The quadratic I interval width was 0.25953, below the prespecified 0.4 precision target. For the 64 control states, I's mean was 0.00367 with interval [-0.00721, 0.01427]; across all 128 states it was -0.01828 [-0.08300, 0.04732]. Ineligible and nonaccepting states contribute their observed no-action outcomes to these summaries. The type contrasts therefore include eligibility and parent-improvement acceptance after the common six-evaluation diagnostic.

\subsection{Initialization, restart and parameters: PC1}
PC1 used the eight JQC1 profiles and $B=1000D$, with three new instances and three seeds per profile and dimension. Nine arms gave 1,296 runs: original JQ100, DE40, DE5D, CMA-native, CMA-restart and four one-factor JQ100 perturbations. DE40 and JQ100 received exactly the first 40 points of the shared uniform master array; DE5D received all $5D$. Both DE policies used synchronous DE/rand/1/bin, $F=0.5$, crossover probability 0.9, reflection and strict replacement. Changing $N$ changes initialization and subsequent population dynamics together.

The CMA arms shared their first native segment: pycma 4.5.0, unit coordinates, initial scale 0.3 and default population. Restart retained the paid global best, drew a new uniform mean with a fresh prescribed stream, and reset CMA state. Only the prespecified function-value, position or stagnation stop-key sets allowed restart; numerical, unknown and other safety stops remained terminal. The cap was 32 restarts; 342 occurred, with at most four in a run. Every new evaluation was charged, including a final partial generation. Native stopping used 994,172 FE and restart used 2,160,000 FE.

The four supplementary configurations changed only $r_0$ or the radius-feedback parameter $\eta$ from JQ100's $0.1\sqrt D$ and 0.7. Default settings of the parameterized copy reproduced the original candidate stream before the formal matrix. All four perturbations were evaluated on every case and retained without selecting a new main configuration.

\begin{table}[!htbp]
\centering\small
\caption{PC1 six-family target-truncated geometric ratios. All comparisons are descriptive. The final two columns retain the worst paired ratio of target-truncated errors in each dimension.}
\label{tab:pc1-details}
\begin{tabular}{lrrrr}
\toprule
Numerator / denominator & 10D & 20D & Worst 10D & Worst 20D \\
\midrule
JQ100 / DE40 & 0.61545 & 0.44457 & $2.00\times10^5$ & $2.00\times10^5$ \\
JQ100 / DE5D & 0.27272 & 0.002412 & 1.077 & 5.528 \\
DE40 / DE5D & 0.44313 & 0.005426 & 1.942 & 5.819 \\
CMA-restart / native & 0.36981 & 0.31661 & 1 & 1 \\
$r_0=0.05\sqrt D$ / JQ100 & 0.90224 & 0.79874 & 337.63 & 4.789 \\
$r_0=0.20\sqrt D$ / JQ100 & 0.85404 & 0.96652 & 10.791 & $2.00\times10^5$ \\
$\eta=0.5$ / JQ100 & 0.83182 & 1.51017 & 15.344 & $2.00\times10^5$ \\
$\eta=0.9$ / JQ100 & 1.05156 & 0.78938 & $2.00\times10^5$ & $2.00\times10^5$ \\
\bottomrule
\end{tabular}
\end{table}

The $\eta$ perturbations reverse macro direction between dimensions. Both tested radii improved the macro mean in both dimensions but retained adverse families and tails; their most adverse family ratios were 1.404/1.161 for $r_0=0.05\sqrt D$ and 1.530/1.204 for $r_0=0.20\sqrt D$ at 10D/20D. Parameter interactions were outside this one-factor design.

\paragraph{Implementation cost.}
A small serial timing study used two repeats per method--dimension condition, with and without trajectory recording. Preparation, recording and background activity affected elapsed time. These measurements characterize the local implementations and do not support a general runtime ranking; solver comparisons here use charged objective evaluations.

\subsection{Public boxed benchmark: BX1}
BX1 used coco-experiment 2.8.3 and the official \texttt{bbob-boxed} suite, whose box is $[-5,5]^D$ and whose infeasible evaluations return positive infinity \citep{cocoboxed}. All 24 functions, official instances 1--3, two seeds and dimensions 10/20 were included. The six prespecified arms were JQ100, DE5D, PSO, Random, L-SHADE and CMA-native; each run had a $1000D$ cap. The adapter supplied raw official objective values, with offsets removed only for the saved error analysis. Initialization, boundary and stop rules remained part of each policy. JQ100/PSO/Random shared the first 40 points, DE used the full $5D$ master, L-SHADE used its native $18D$ initialization, and CMA used its native population and a separately generated mean.

The same two search-seed identifiers were reused across functions and instances for all methods. Master initial populations and CMA means vary by problem and seed; L-SHADE has only two distinct initial coordinate arrays per dimension. Thus the 288 master arrays do not imply 288 independent algorithmic starts, and shared seeds may induce cross-function dependence.

\begin{table}[!htbp]
\centering\small
\caption{BX1 raw endpoint wins/ties/losses (W/T/L) for JQ100 and worst target-truncated pair ratios. Each dimension has 144 paired cases per reference. Raw ordering and the primary $\max(e,0.01)$ ratio measure different precision levels.}
\label{tab:bx1-tails}
\begin{tabular}{llrr}
\toprule
$D$ & Reference & Raw W/T/L & Worst primary pair \\
\midrule
10 & DE5D & 130/0/14 & 180.302 \\
10 & PSO & 95/0/49 & 7.847 \\
10 & Random & 144/0/0 & 0.671 \\
10 & L-SHADE & 53/2/89 & 191.343 \\
10 & CMA-ES & 45/1/98 & 610.384 \\
20 & DE5D & 136/0/8 & 9.347 \\
20 & PSO & 91/2/51 & 316.981 \\
20 & Random & 144/0/0 & 0.515 \\
20 & L-SHADE & 63/0/81 & 34260.1 \\
20 & CMA-ES & 30/4/110 & 387004 \\
\bottomrule
\end{tabular}
\end{table}

CMA-native stopped early in 119/144 10D runs and 113/144 20D runs, spending 812,770 and 1,778,504 FE respectively. Each other method spent 1,440,000 and 2,880,000 FE. Target levels were $1,10^{-2},10^{-4},10^{-8}$. The original curves used completed batches; retrospective curves use ordered individual values at $0.1,0.25,0.5,0.75,1$ times the cap. Reconstructing the running minimum reproduces every recorded batch best and terminal result. Nine of 240 target--budget cells gain one to three hits at intermediate budgets; all terminal hit counts are unchanged. The macro summaries weight the 24 functions equally and apply to this fixed set of problems and initialization rules.

\section{Statistical hierarchy and saved-data sensitivity}\label{app:followup-statistics}

CT1 used 4,000 percentile bootstrap resamples (seed 23109999), drawing the 16 instances within each profile--dimension cell and then two paired futures within each sampled state. Type and radius contrasts shared the same resamples. Its primary quadratic group contains 64 states across four cells; cells and states within cells receive equal weight.

S1 and BX1 each used 2,000 resamples (seeds 2026102901 and 23500999). S1 sampled instances within fixed profiles and then paired search seeds or futures, preserving the original family weighting. The original BX1 analysis sampled three instances within each fixed function and then paired seeds independently by function.

The retrospective analysis instead draws the two search-seed labels jointly across every function and instance, then draws three instances within each function with that instance draw shared across sampled seeds. All solver comparisons remain paired. Its 2,000 replicates preserve cross-function seed dependence. The mean ratios and their directions remain unchanged; the two observed seed blocks limit the empirical variation represented by the intervals.

The CT1 and S1 95\% intervals are pointwise; the fixed functions are not resampled. Future streams within a state and search seeds within an instance are repeated observations at lower levels. PC1 originally reported descriptive aggregation; its matched-start contrasts now include retrospective intervals, with the construction described below.

The retrospective PC1 analysis uses the same crossed construction with three shared search seeds, three instances within each profile and six-family equal weights. It uses 2,000 replicates. For JQ100/DE40, intervals from the alternative nested instance-then-seed resampling are [0.459, 1.150] at 10D and [0.146, 1.394] at 20D, also spanning parity. Table~\ref{tab:pc1-families} reports every family's mean ratio and target hits. With only three empirical seed blocks, these intervals describe design sensitivity rather than precise population coverage.

PC1 uses three search-seed labels across all profiles and instances; BX1 uses two across all functions and instances. In BX1, the shared master has $5D$ points and is generated separately for each problem--seed pair; L-SHADE reuses two native initial coordinate arrays per dimension. CT1 uses independently generated state instances with two paired futures per state.

S1's five analyses covered endpoint distributions and paired wins/losses, target arrival and ECDFs, error representation, threshold sensitivity, and hierarchical hit-rate intervals. Its 2,026 saved trajectories comprise 960 JQC1, 450 JQX1, 384 J6 and 232 M3 trajectories, with 77 selected intervention states and 308 paired futures. These are retrospective analyses; the old primary metrics and decisions remain identifiable.

For JQX1 and the original batch summaries, the first visible target hit is the end of its completed evaluation batch; its first within-batch occurrence lies after the preceding batch end and at or before that recorded endpoint. Native nonhits are censored at actual termination FE, and carrying the paid best to the cap creates neither evaluations nor new hits.

\begin{table}[!htbp]
\centering\small
\caption{S1 retrospective JQX1 quadratic error-representation sensitivity, JQ100/reference. Floored errors use $\max(f,\epsilon)$; target truncation uses each profile's original target.}
\label{tab:s1-scales}
\begin{tabular}{llrrrr}
\toprule
$D$ & Reference & $\epsilon=10^{-15}$ & $10^{-12}$ & $10^{-9}$ & Target clip \\
\midrule
10 & S0 & 548.776 & 104.470 & 3.304 & 0.405 \\
10 & CMA-ES & $2.503\times10^9$ & $4.142\times10^7$ & 41424.8 & 2.165 \\
20 & S0 & 0.376 & 0.376 & 0.376 & 0.661 \\
20 & CMA-ES & $1.552\times10^9$ & $9.080\times10^7$ & 618615 & 33.380 \\
\bottomrule
\end{tabular}
\end{table}

The 10D comparison to S0 reverses between floored precision and target truncation, whereas the comparison to CMA-ES retains its direction. Reexpressing saved values with shifts $-1,0,1$ and subtracting the shifted optimum recovered the original values in exact decimal arithmetic; floating-point cancellation reached $7.11\times10^{-15}$, with no changed target classifications in these records. This checks saved-value representation, not a newly executed shifted-objective policy.

In the prespecified sensitivity list for the retrospective analysis, JQC1's macro screens stayed unchanged under a $\pm20\%$ variation of their ratio threshold. JQX1 10D/NQ passed the original worst-pair cap 10 but failed cap 5; 20D/S0 failed cap 10 but passed cap 20. The original worst pairs were 5.259 and 13.839. The original decisions remain attached to their original caps.

For hit-rate uncertainty, all-zero or all-one empirical strata can yield bootstrap intervals [0,0] or [1,1]; these describe the finite empirical resampling distribution rather than certifying a population probability. Sparse M3 selected-state exposure also remains in its denominator and interval conditions.

\section{Sampling and corrected observations}\label{app:followup-integrity}

CT1 used newly generated instances. Twelve numerical seed identifiers overlapped with identifiers used for M3 continuations, but the resulting random streams and actual problem and initialization arrays were distinct. CT1 is analysed as its own preselected population and is not pooled with J6 or M3.

In BX1, the stored initial objective values for all 288 runs of each of JQ100, DE and PSO were inadvertently overwritten after evaluation. The original initial points were reevaluated with the same official objective implementation to recover these 864 initial batches. Subsequent observations, historical-best values recorded during search and terminal outcomes were unchanged. The corrected values were used in the saved-trajectory analysis; no configuration was changed or selected on that basis.
\FloatBarrier

\FloatBarrier
\bibliographystyle{plainnat}
\bibliography{references}
\end{document}